\documentclass[letterpaper, 10 pt, conference]{ieeeconf}  % Comment this line out if you need a4paper

\IEEEoverridecommandlockouts                   
\usepackage[utf8]{inputenc} % allow utf-8 input
\usepackage[T1]{fontenc}    % use 8-bit T1 fonts
\usepackage{hyperref}       % hyperlinks
\usepackage{url}            % simple URL typesetting
\usepackage{booktabs}       % professional-quality tables
\usepackage{amsfonts}       % blackboard math symbols
\usepackage{nicefrac}       % compact symbols for 1/2, etc.
\usepackage{microtype}      % microtypography
\usepackage[table]{xcolor}% colors

\usepackage{graphicx}
\usepackage{amsmath}
\usepackage{amssymb}

\usepackage{listings}
\usepackage{mwe} % for placeholder images
\usepackage{makecell}
\usepackage[normalem]{ulem} % strikethrough font
\usepackage{algorithm}
\usepackage{algorithmic}

\usepackage{wrapfig}
\usepackage{caption}
\usepackage{pifont} % Checkmark and Crossmark
\usepackage{multirow}
\usepackage{float}

\usepackage{chngpage}
\usepackage{adjustbox}
\usepackage{comment}
\usepackage{bm}
\usepackage[capitalize]{cleveref}

\usepackage{mathtools}

\usepackage{array}

\crefname{section}{Sec.}{Secs.}
\Crefname{section}{Section}{Sections}
\Crefname{table}{Table}{Tables}
\crefname{table}{Tab.}{Tabs.}
\definecolor{Gray}{gray}{0.9}
\definecolor{ImportantColor}{rgb}{0.63, 0.79, 0.95}
\newcolumntype{g}{>{\columncolor{ImportantColor}}c}
\newcolumntype{?}{!{\vrule width 1pt}}

\definecolor{tarlblue}{HTML}{1F77B4}
\definecolor{tarlpink}{HTML}{E377C2}

\def\imw#1#2{\includegraphics[width=#2\linewidth]{#1.png}}
\def\imwjpg#1#2{\includegraphics[width=#2\linewidth]{#1.jpg}}

\newcommand{\tb}[3]{\setlength{\tabcolsep}{#2mm}\begin{tabular}{#1}#3\end{tabular}}

\graphicspath{
{figures/},
}

\newcommand*{\belowrulesepcolor}[1]{% 
  \noalign{% 
    \kern-\belowrulesep 
    \begingroup 
      \color{#1}% 
      \hrule height\belowrulesep 
    \endgroup 
  }%
} 
\newcommand*{\aboverulesepcolor}[1]{% 
  \noalign{% 
    \begingroup 
      \color{#1}% 
      \hrule height\aboverulesep 
    \endgroup 
    \kern-\aboverulesep 
  }%
}

\definecolor{codegreen}{rgb}{0,0.6,0}
\definecolor{codegray}{rgb}{0.5,0.5,0.5}
\definecolor{codepurple}{rgb}{0.58,0,0.82}
\definecolor{backcolour}{rgb}{0.95,0.95,0.92}

\lstdefinestyle{mystyle}{
    backgroundcolor=\color{backcolour},   
    commentstyle=\color{codegreen},
    keywordstyle=\color{magenta},
    numberstyle=\tiny\color{codegray},
    stringstyle=\color{codepurple},
    basicstyle=\ttfamily\footnotesize,
    breakatwhitespace=false,         
    breaklines=true,                 
    captionpos=b,                    
    keepspaces=true,                 
    numbers=left,                    
    numbersep=5pt,                  
    showspaces=false,                
    showstringspaces=false,
    showtabs=false,                  
    tabsize=2
}

\usepackage{wrapfig}   % preamble
\usepackage{calc}   % 放在 preamble
\usepackage{arydshln}   % preamble
\providecommand{\iflatexml}{\iffalse}% \input{math_commands.tex}

\newcommand{\method}{TaRL}
\newcommand{\methodfull}{\underline{Ta}ctile \underline{R}eward \underline{L}earning}
\title{\method{}: Learning General and Physical Rewards from Tactile Demonstrations}

\title{\LARGE \bf
\method{}: Learning General and Physical Rewards from Tactile Demonstrations
}

\iflatexml
\author{Po-Yi Wu (National Taiwan University) \tt b11902127@csie.ntu.edu.tw
\and Dao-Jan Chang (National Taiwan University) \tt r15922154@csie.ntu.edu.tw
\and Shang-Ya Hsiao (Delta Electronics) \tt angelica.hsiao@deltaww.com
\and Hong-Ming Chen (Delta Electronics) \tt hongming.chen@deltaww.com
\and Yu-Cheng Su (Delta Electronics) \tt yucheng.su@deltaww.com
\and Tsung-Wei Ke (National Taiwan University) \tt twke@csie.ntu.edu.tw}
\else
\author{Po-Yi Wu$^{1}$, Dao-Jan Chang$^{1}$, Shang-Ya Hsiao$^{2}$, Hong-Ming Chen$^{2}$, Yu-Cheng Su$^{2}$, and Tsung-Wei Ke$^{1}$\\
\vspace{2pt}
{\normalsize $^{1}$National Taiwan University \quad $^{2}$Delta Electronics}\\
{\tt\small \{b11902127, r15922154, twke\}@csie.ntu.edu.tw}\\
{\tt\small \{angelica.hsiao, hongming.chen, yucheng.su\}@deltaww.com}
}
\fi

\begin{document}

% \maketitle
% \thispagestyle{empty}
% \pagestyle{empty}

\twocolumn[{
\renewcommand\twocolumn[1][]{#1}%
\maketitle
\thispagestyle{empty}
\pagestyle{empty}
\begin{center}
\newcommand{\teaserwidth}{\textwidth}
\vspace{-0.15in}
\centerline{
\includegraphics[width=\teaserwidth, clip]{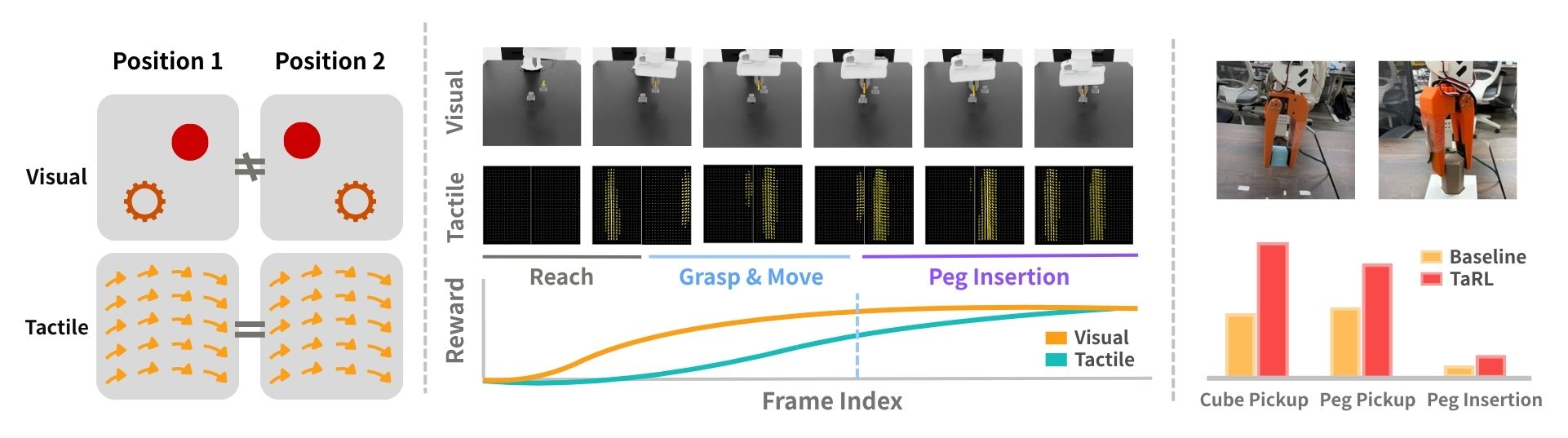}
}
\captionof{figure}{
\textbf{\method{} is the first approach to learning RL rewards from tactile demonstrations.}
\textbf{Left:} Tactile observations are invariant to varying object positions.  \textbf{Middle: } Tactile rewards capture local robot-object interaction, which cannot be observed obviously.  This allows \method{} to produce informative feedback at different contact phases, which visual reward learning approaches cannot offer.  \textbf{Right:} \method{} enhances RL efficiency on contact-rich manipulation in the real world.
% \method{} produces reward signals from tactile observations to facilitate RL in robotic manipulation.  \textbf{Left:} tactile observations are invariant to varying object positions.  \textbf{Middle: } tactile rewards capture local robot-object interaction, which cannot be observed visually.  This allows \method{} to produce informative feedback at different contact phases, which visual reward learning approaches cannot offer.  \textbf{Right:} \method{} enhances RL efficiency on contact-rich manipulation in the real world.
}
\label{fig:teaser}
\end{center}
}]

\begin{abstract}
Contact-rich manipulation requires robots to sequence precise contacts, maintain stable grasps, and apply directed forces. Reinforcement learning (RL) can acquire such behaviors automatically, but its performance hinges on reward design: sparse rewards reduce the learning efficiency, while dense rewards are hard to specify. Visual reward learning addresses this by inferring rewards from action-free demonstrations. Because it conditions only on visual observations, it fails to capture rewards beyond visual goals. We propose \methodfull{} (\method{}), a framework that learns rewards from tactile demonstrations.  \method{} takes a sequence of tactile deformation maps as input, and regresses task-completion progress from both successful and failed demonstrations. Because \method{} captures local robot-object interaction, it provides informative feedback to learn firm grasps and correctly directed forces; meanwhile, it is robust to changes in scene layout such as object position. We evaluate \method{} on four manipulation tasks in simulation and two in the real world. Used as a shaping reward, it substantially improves both sample efficiency and final success rate, raising success on Nut threading from 34\% to 56\% in simulation and on cube pickup from 37\% to 97\% in the real world. Combining tactile with visual rewards improves performance further. \method{} also generalizes across object instances: trained on \textit{box placement} and directly deployed to \textit{can placement}, it significantly improves policy learning on the new task. Project page is available at \url{https://embodiedai-ntu.github.io/tarl}.
\end{abstract}

\section{Introduction}

Robotic manipulation features contact-rich interactions between robots and objects. Manipulation tasks like tool use and assembly require the robot to orchestrate sequences of precise contacts, stable grasps, and directed forces~\cite{manipulation}. Reinforcement learning offers a general framework for acquiring such behaviors by optimizing policies against task rewards~\cite{rajeswaran2017learning}; however, its performance is bottlenecked by reward design: sparse rewards make learning inefficient, while dense rewards are difficult to specify and may induce unintended behavior without careful inspection~\cite{amodei2016concrete}.

Reward learning from demonstrations (RLfD) aims to address such limitations~\cite{ng2000algorithms,abbeel2004apprenticeship}: it extracts underlying rewards from demonstrations, and produces dense, informative feedback to facilitate downstream RL~\cite{finn2016connection}.
Recently, \textit{visual reward learning} has emerged as a dominant RLfD paradigm in robotic manipulation~\cite{finn2016guided,fu2018variational}.  This approach represents rewards as visual goals depicted in demonstrations, such as the desired object movements or a robot's reaching positions and orientations~\cite{zhang2025rewind}.  Because visual goals can be captured without action annotations and can be transferred from other expert embodiments like humans~\cite{chen2021learning,zakka2022xirl,alakuijala2023learning,kumar2023graph}, visual reward learning has shown promising scalability with large-scale human video data~\cite{liang2026robometer}.

While attracting great interest, visual reward learning has limited expression, incapable of capturing all desired outcomes of contact-rich manipulation.  Take assembly or insertion tasks as examples, both involve local robot-object interactions describing whether a grasp is firm, how to apply forces, and when to switch movement patterns. Because these goals are occluded from view, they are overlooked by visual rewards. In contrast, tactile sensing~\cite{dahiya2009tactile} directly encodes such physical information, which motivates us to study whether tactile perception yields reward functions and whether combining it with visual rewards enhances RL on contact-rich manipulation tasks.
% While attracting great interest, visual reward learning is bounded by what a camera can see, and cannot capture all desired outcomes of contact-rich manipulation. Take assembly or insertion as examples. Both hinge on local robot-object interactions, namely whether a grasp is firm, how force is applied, and when to switch movement patterns. These states are occluded by the gripper and produce no change in appearance, so a visual reward sees the outcome but not the process and degrades to a sparse signal exactly when dense feedback matters most (Fig.~\ref{fig:teaser}). Tactile sensing~\cite{dahiya2009tactile} directly encodes this physical information. We therefore ask whether tactile perception yields reward functions, and whether combining it with visual rewards enhances RL on contact-rich manipulation tasks.

We introduce \methodfull{} (\method{}), to the best of our knowledge, the first RLfD approach that learns RL reward functions from tactile demonstrations of contact-rich manipulation.  To isolate the effect of the tactile modality, we build \method{} upon a common reward learning framework~\cite{zhang2025rewind} and leave more advanced design as future work.  Using both successful and failed demonstrations~\cite{shiarlis2016inverse}, \method{} conditions on a sequence of tactile deformation maps and regresses task completion progress, without access to ground-truth states~\cite{kwon2023reward}, visual observations~\cite{wang2024rl}, or action annotations~\cite{ziebart2008maximum}. Once pre-trained offline, \method{} can be plugged directly into downstream RL, providing dense reward signals to guide policy exploration without online fine-tuning. \method{} has two key properties: (1) it pursues physical goals that visual observations cannot capture, and is therefore complementary to visual reward learning; and (2) it generalizes across object positions and scene layouts, since tactile signals encode local contact rather than global scene appearance (Fig.~\ref{fig:teaser}).

We evaluate \method{} on a wide range of robotic manipulation tasks in both simulation and the real world, where it consistently improves downstream RL training efficiency. In simulation, we test four tasks--\textit{pick and place}, \textit{peg insertion}, \textit{gear assembly}, and \textit{nut threading}; in the real world, we test \textit{pick and place} and \textit{peg insertion}. Across all experiments, \method{} is trained on tactile demonstrations collected from a single object position, while downstream RL policies are evaluated on a disjoint set of positions. Empirical results show that \method{} improves RL training efficiency while generalizing to unseen object instances in pick-and-place.

We summarize our contributions as follows: (1) we propose tactile reward learning, a new RLfD paradigm orthogonal to existing visual reward learning; (2) we evaluate \method{} on diverse manipulation tasks in both simulation and the real world, where it raises final success by up to 22\% in simulation and 60\% in the real world; (3) \method{} exhibits strong generalization to novel object positions and instances; (4) combining \method{} with visual reward learning produces more complete reward signals which significantly enhance RL's training efficiency.

\section{Related Work}

\textbf{Reward learning from demonstrations.}
RLfD recovers task objectives from expert behavior in place of hand-engineered rewards. Inverse RL~\cite{ng2000algorithms, abbeel2004apprenticeship, ziebart2008maximum, fu2017learning} typically assumes action-labeled trajectories, whereas observation-only approaches can learn from demonstrations collected under different embodiments~\cite{chen2021learning, zakka2022xirl, alakuijala2023learning, kumar2023graph}. Visual reward learning has been particularly well studied, with rewards derived from trajectory preferences~\cite{wang2024rl, yang2024rank2reward}, motion tracking~\cite{xiong2021learning, hsieh2025dexman}, success classification~\cite{fu2018variational}, and task-progress representations~\cite{ma2022vip, TCN2018, zhang2025rewind, liu2025timerewarder}, increasingly using vision-language and video foundation models~\cite{sontakke2023roboclip, rocamonde2024vision, ma2025vision, liang2026robometer}. Despite their different formulations, these methods ground reward in scene appearance, which leaves contact-centric variables such as grasp stability and applied force largely unobserved.

\textbf{Tactile sensing in robot learning.}
High-resolution tactile sensors~\cite{yuan2017gelsight} and tactile simulators~\cite{chen2024general, akinola2025tacsl} expose local contact geometry, pressure, force, and slip that are difficult to infer visually. Prior work primarily uses tactile observations as policy inputs for in-hand rotation~\cite{touchdexterity2023}, assembly~\cite{lin2023bi}, and force-aware diffusion policies~\cite{helmut2025tactile}, or as representations for prediction and planning~\cite{ai2024robopack}. In contrast, tactile observations have received considerably less attention as a source for learning the \emph{reward itself}. See-to-Touch~\cite{guzey2024see}, for example, learns tactile-conditioned policies but derives reward visually through optimal transport to an expert video. TaRL instead learns a reward representation from tactile demonstrations. 
\section{Method}
\label{sec:method}

\begin{figure*}[t]
    \centering
    \begin{adjustbox}{width=\linewidth}
    \tb{@{}c@{}}{0.1}{
    \imw{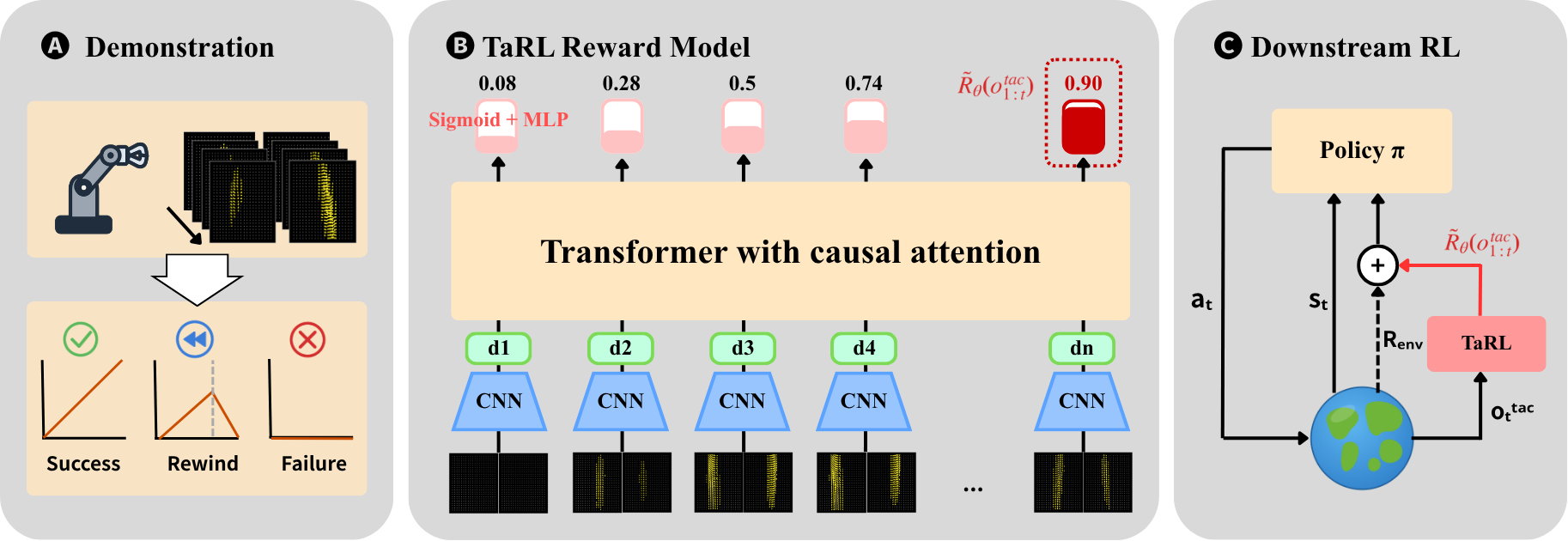}{0.92}
    }
    \end{adjustbox}
    \caption{
    \textbf{Overview of \method{}.}  \textbf{Left:} It is trained with successful, rewound and failed demonstrations. \textbf{Middle:} It conditions on tactile deformation maps and outputs task-progress rewards. \textbf{Right:} It is applied to downstream RL, offering dense rewards.
    %\textbf{Overview of \method{}.} \method{} is the first approach to learning rewards from tactile demonstrations. \textbf{Left:} It is trained with successful, rewound and failed demonstrations. \textbf{Middle:} It conditions on a sequence of tactile deformation maps and regresses task-progress rewards. \textbf{Right:} Once trained, it can be plugged into downstream RL, offering dense rewards.
    }
    \label{fig:method}
    \vspace{-16pt}
\end{figure*}

\method{} learns to extract reward functions from tactile demonstrations to enhance RL-based policy learning in contact-rich manipulation.  It models rewards as task completion progress, a parameterization that matches how tactile signals evolve through successful manipulation---from initial contact to stable grasp to directed force application---providing dense, precise feedback at every step.  Because tactile rewards do not depend on scene appearance, they are robust to changes in scene layout and can even transfer across tasks. Fig.~\ref{fig:method} illustrates the \method{} pipeline. Sec.~\ref{subsec:problem_formulation} formalizes tactile reward learning; Sec.~\ref{subsec:model} describes \method{}'s input modality, data collection and preprocessing, architectural design and training objectives; Sec.~\ref{subsec:reward_shaping} presents the deployment of \method{} to downstream RL.

\subsection{Problem Formulation}
\label{subsec:problem_formulation}

A robotic manipulation task defines a finite-horizon Markov decision process (MDP) $\mathcal{M} = (\mathcal{S}, \mathcal{O}, \mathcal{A}, P, R, \rho, \gamma)$, where $\mathcal{S}$ is the state space, $\mathcal{O}$ is the observation space, $\mathcal{A}$ is the action space, $P$ denotes the transition dynamics of the environment, $R$ is the reward function of the task, $\rho$ is the initial state distribution, and $\gamma$ denotes the discount factor.  RL seeks the optimal policy $\pi^*$ that maximizes the expected discounted return over time horizon $H$: $\pi^* = \arg \max_\pi \mathbb{E}_{\tau}\left[ \sum_{t=0}^{H-1} \gamma^t R(s_t, a_t, s_{t+1}) \right]$, where $\tau=(s_0,a_0,\dots,s_T)$ is a trajectory induced by the policy and environment dynamics. In practice, the ground-truth dense reward $R$ is unknown, while sparser rewards, indicating success or failure of the entire task or at each manipulation stage, provide insufficient supervision.  Common practice therefore resorts to laborious reward engineering~\cite{ng1999policy}, handcrafting auxiliary reward functions to guide policy learning.

\paragraph{Visual reward learning}
Prior work has explored modeling visual rewards in several forms, including trajectory preferences~\cite{wang2024rl}, motion tracking~\cite{hsieh2025dexman}, task success classification~\cite{fu2018variational}, and task progress regression~\cite{TCN2018}.  Because our goal is to highlight the novelty of learning rewards from tactile observations, we focus on the most common approach---progress regression, in this work and leave the study on other forms of visual rewards as future work.

Given a goal image $g$, one common formulation defines the reward as the change in goal-embedding distance between consecutive observations~\cite{ma2022vip}:
$\hat{R}(o_t, o_{t+1}; \phi, g) := D_\phi(o_{t+1}; g) - D_\phi(o_t; g),$
where $o_t$ is the observation at state $s_t$ and $D_\phi$ is a learned metric parameterized by $\phi$.  This reward encourages progress toward the goal and can be plugged directly into the RL objective\footnote{Integrating observation-based rewards into the RL objective is straightforward under a deterministic emission probability $p(o_t|s_t)$}.  However, it has two limitations: (1) it requires a goal image at inference time, which is often unavailable, and (2) it measures progress between adjacent frames, where the embedding change is often too small to provide informative signal.
ReWiND~\cite{zhang2025rewind} addresses both limitations with a more general formulation. Instead of conditioning on a goal image, it conditions on the trajectory observed so far and predicts task progress directly, leaving the model to infer the goal implicitly from the task instruction. The model is trained to regress the following target: $\hat{R}_\phi(o_{1:t})=\frac{t}{T}$, with time horizon $T$.
% \begin{align}
%     R_v(o_{1:t}) =
%     \begin{cases} 
%         \frac{t}{T} & \text{if } o_{1:t} \text{ sampled from successful trajectories} \\
%         0 & \text{otherwise} \\
%     \end{cases}
% \end{align}

\subsection{Tactile Reward Learning}
\label{subsec:model}
Since our goal is to study tactile feedback as alternative reward signals rather than innovating model design, we adopt the same trajectory-conditioned formulation and model architecture as ReWiND~\cite{zhang2025rewind}, but replace RGB videos with tactile videos as the conditioning input. For completeness, we briefly describe its input modality, data collection and preprocessing, model architecture, and training objectives.

\paragraph{Demonstration collection}
Training \method{} requires both successful and failed demonstrations. In simulation, we train robot policies via RL for data collection.  We collect rollouts from intermediate checkpoints of an RL training run, yielding a mix of failures, near-misses, and successes with diverse robot-object interactions.  In the real world, we collect demonstrations via manual teleoperation, including both successful and failed attempts.  For each rollout, we record the binary task outcome and the full tactile observation sequence; these form the training dataset for \method{}.  %Check the Appendix for more data collection details.

\paragraph{Tactile observations}
Tactile sensors come in a range of designs offering different modalities, including binary contact signals~\cite{touchdexterity2023}, discrete 2D orientations~\cite{qi2023general}, near-surface depth maps~\cite{yuan2017gelsight}, and 3D deformation maps~\cite{chen2024general,si2024difftactile}. Among these, 3D deformation maps most directly capture robot-object physical interactions, such as the pressure and shear forces applied at the contact surface, and therefore provide the richest feedback for contact-rich manipulation.  We adopt 3D deformation maps in \method{}. Our setup is a single-arm robot equipped with a two-finger gripper, with one tactile sensor mounted on each finger. At each timestep $t$, we obtain a pair of deformation maps $o^{tac}_t = (o^{tac}_{t, 1}, o^{tac}_{t, 2})$, where $o^{tac}_{t, i} \in \mathbb{R}^{3 \times H \times W}$ denotes the deformation map of finger $i$, and $H, W$ are the spatial resolution.

\paragraph{Model architecture}
\method{} conditions on the 3D deformation maps observed so far and predicts task progress. From the variable-length trajectory of past observations $o^{tac}_{1:t}$, we uniformly subsample a fixed number of timesteps ($N=16$); if fewer than $N$ timesteps have been observed ($t < N$), we pad with the most recent frame. At each sampled timestep, the two deformation maps (one per finger) are independently encoded by a shared 3-layer convolutional network into $c$-dimensional features, which are then concatenated to form a $2c$-dimensional tactile token $d$.  The resulting sequence $(d_1, \dots, d_N)$ is processed by a transformer~\cite{vaswani2017attention} with causal attention, so that each output token depends only on its own and earlier inputs---matching the causal structure of task progress, which at any timestep depends only on what has happened so far. A shared MLP with a sigmoid output $\sigma$ is then applied to each output token to produce per-timestep progress predictions.  Denoting all model parameters by $\theta$, the tactile reward at timestep $t$ is the task progress predicted from the most recent sampled token: $\tilde{R}_\theta(o^{tac}_{1:t}) = \sigma \bigl( \mathrm{MLP} (d_N) \bigr)$.  Meanwhile, owing to causal attention, the progress predicted from each of intermediate tokens $d_{k}$ with $k<N$ is equivalent to the reward at the corresponding timestep $t_k = \lfloor \frac{t \times k}{N} \rfloor$: $\tilde{R}_\theta(o^{tac}_{1:t_k}) = \sigma \bigl( \mathrm{MLP} (d_k) \bigr)$.

\paragraph{Data preprocessing}
Before feeding the $N$ sampled deformation maps into \method{}, we normalize them per-sequence so that the reward model learns from relative rather than absolute surface displacements.  Specifically, we divide every map in the sequence by the maximum absolute entry across all $N$ maps; this scaling factor is clipped to a minimum of  $10^{-6}$ to handle pre-contact frames where the maximum is near zero. By removing absolute force magnitudes from the input, this normalization encourages the model to infer rewards from the spatio-temporal patterns of robot-object interaction rather than from specific deformation values. As a result, the learned reward generalizes across episodes and object instances, and even transfers across tasks, where absolute deformation values can differ substantially.

\paragraph{Training Objective}
We apply a similar data augmentation strategy as ReWiND that constructs three sets of training examples from the demonstrations: (1) original tactile videos $o^{tac}_{1:t}$ from successful demonstrations, with target $t/T$; (2) rewound tactile videos from successful demonstrations, formed by concatenating a forward prefix with a reversed suffix at a random split point $i$, with target $(i-t)/T$; and (3) original tactile videos from failed demonstrations, with target $0$.  The reward regression loss is:
\begin{align}
    &\mathcal{L}(\theta) = \underbrace{\sum_{t=1}^T \bigl( \tilde{R}_\theta(o^{tac}_{1:t}) \bigr)^2}_{\text{failed tactile videos}} + \underbrace{\sum_{t=1}^T \bigl( \tilde{R}_\theta(o^{tac}_{1:t}) - \frac{t}{T} \bigr)^2}_{\text{successful tactile videos}} \nonumber \\
    &+ \underbrace{\mathbb{E}_{\substack{i \sim \mathcal{U}(2, T-1) \\ t \sim \mathcal{U}(1,i-1)}} \bigl[ \bigl( \tilde{R}_\theta([o^{tac}_{1:i}; o^{tac}_{i-1:i-t}]) - \frac{i-t}{T} \bigr)^2 \bigr]}_{\text{successful but rewound tactile videos}},
\end{align}
where $\mathcal{U}(1, t)$ denotes uniformly sampling an integer from $[1, t]$.  
% See the Appendix for full details.

\subsection{Downstream RL with \method{}}
\label{subsec:reward_shaping}

After offline training, \method{} is plugged directly into downstream RL to provide dense shaping rewards during rollout collection, without further fine-tuning. At each timestep $t$, we uniformly subsample $N$ tactile deformation maps from the tactile video observed so far in the current episode, $o^{tac}_{1:t}$, padding with the most recent frame if fewer than $N$ timesteps have been observed. \method{} then predicts the tactile reward $\tilde{R}_\theta(o^{tac}_{1:t})$, which we combine with the environment's task reward $R(s_t, a_t, s_{t+1})$ and visual rewards $\hat{R}_\phi(o_{1:t})$ predicted by ReWiND to form the total reward signal:$R^{\text{total}}(s_t, a_t, s_{t+1}) = R(s_t, a_t, s_{t+1}) + \alpha\,\tilde{R}_\theta(o^{tac}_{1:t}) + \beta\,\hat{R}_\phi(o_{1:t})$, where $\alpha, \beta$ are hyperparameters controlling the weight of the tactile and visual shaping term.  Notably, the environment's task reward $R(s_t, a_t, s_{t+1})$ does not contain terms that encourage stable grasp or precise contact, because our goal is to verify whether such structure can be learned automatically from tactile demonstrations. The policy is then optimized with off-the-shelf RL algorithms.  In our simulation experiments, we conduct online RL with Proximal Policy Optimization (PPO)~\cite{schulman2017proximal}; in our real-world experiments, we perform offline RL with Implicit Q-Learning (IQL)~\cite{kostrikov2021offline}.  See Sec.~\ref{sec:experiment} for training details.
\begin{figure}[ht]
  \vspace{-5pt}
  \centering
  \begin{adjustbox}{width=\linewidth}
    \tb{@{}cc@{\hspace{6pt}}!{\vrule width 0.6pt}@{\hspace{6pt}}c@{}}{0.2}{
      \multicolumn{2}{c}{\textbf{Simulation}} & \textbf{Real world} \\ [3pt]
      \imwjpg{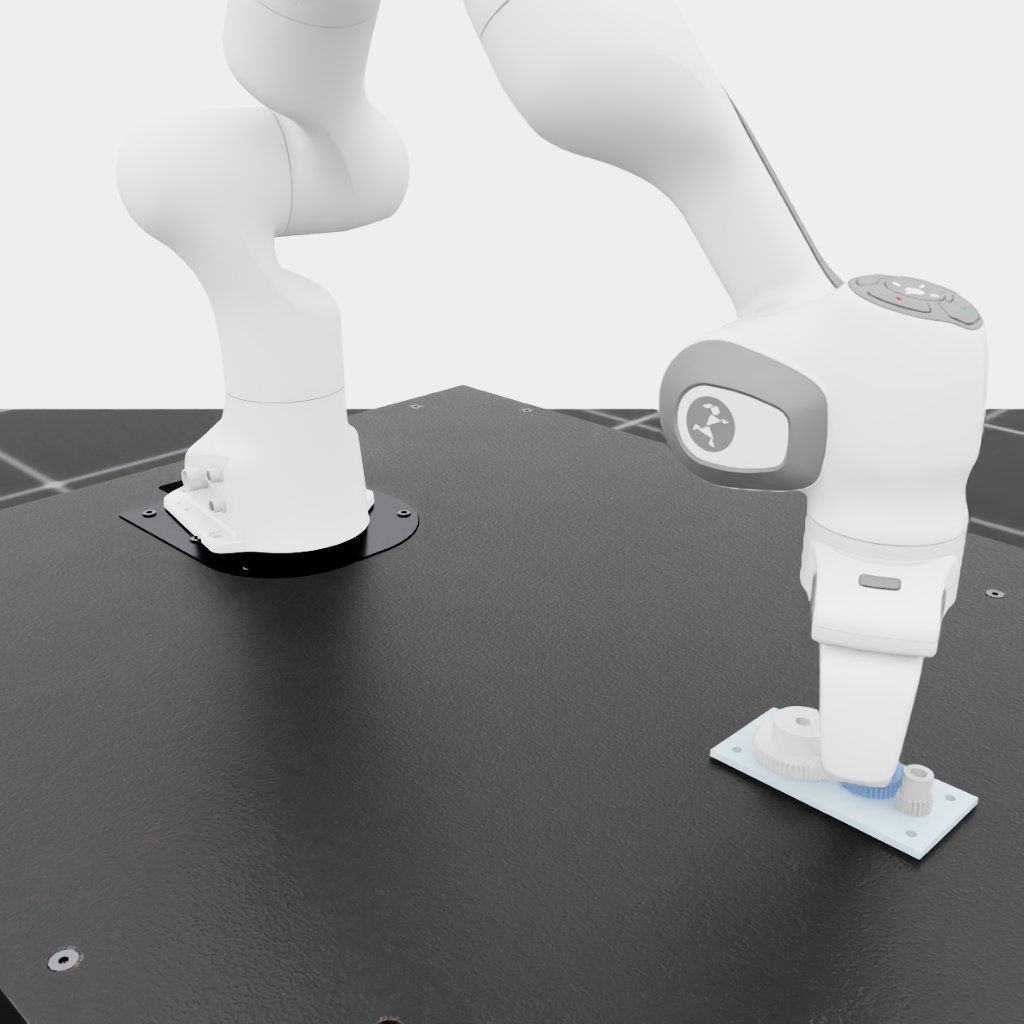}{0.33} &
      \imwjpg{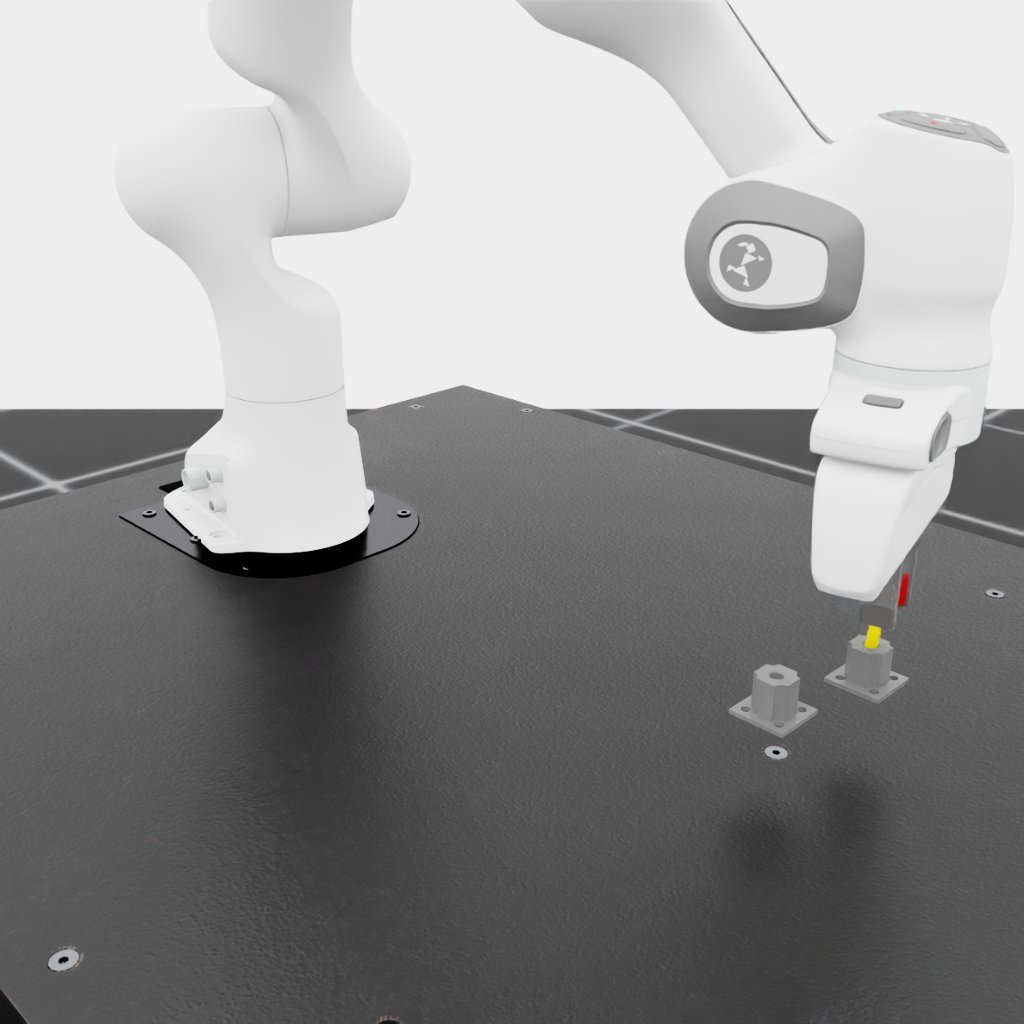}{0.33} &
      \imwjpg{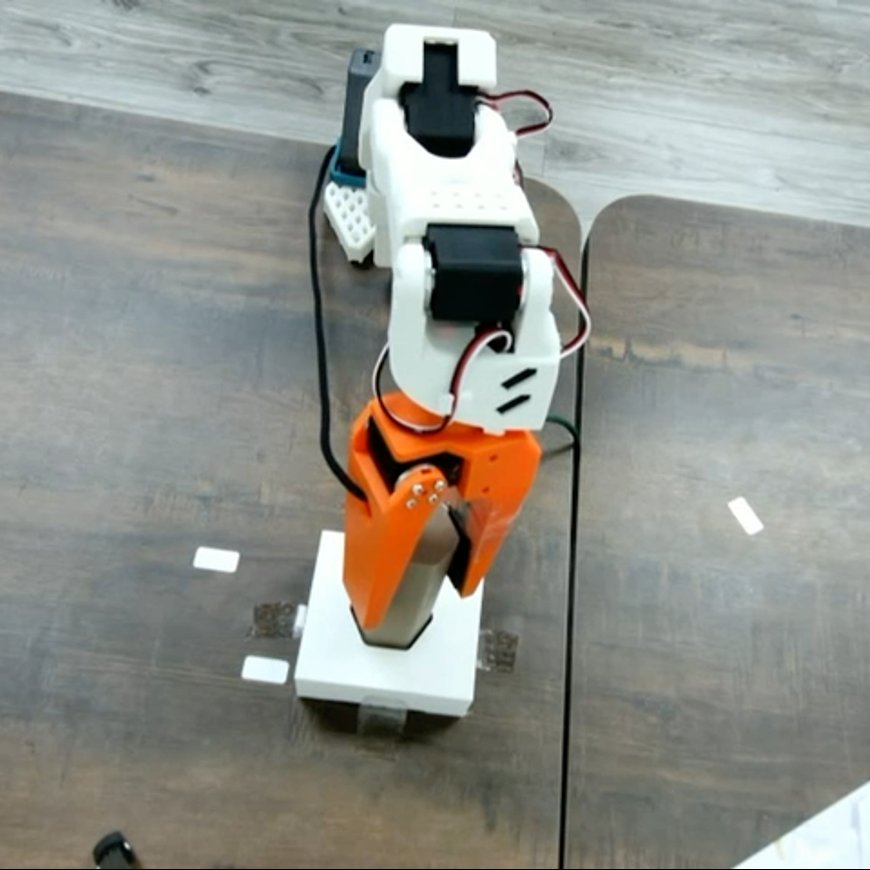}{0.33} \\
      (a) Gear assembly & (b) Peg insertion & (e) Peg insertion \\ [4pt]
      \imwjpg{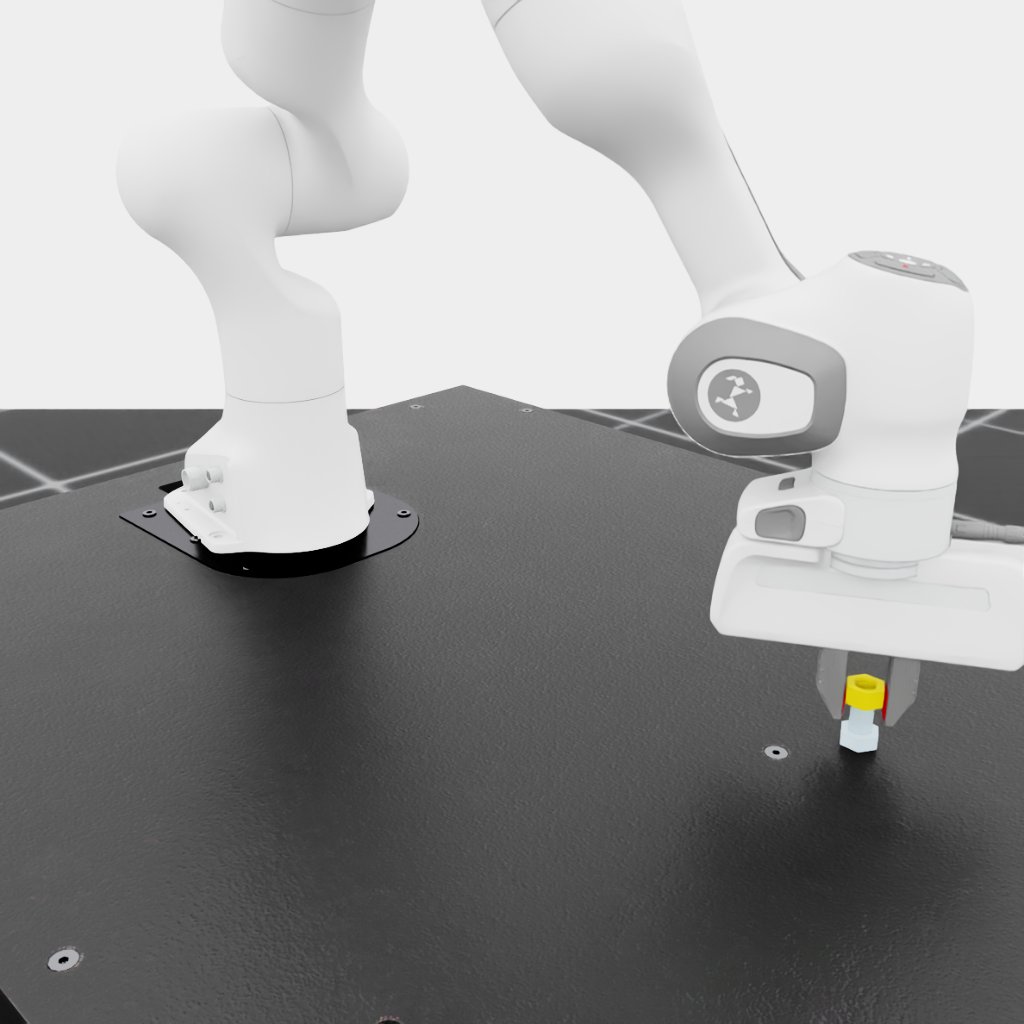}{0.33} &
      \imwjpg{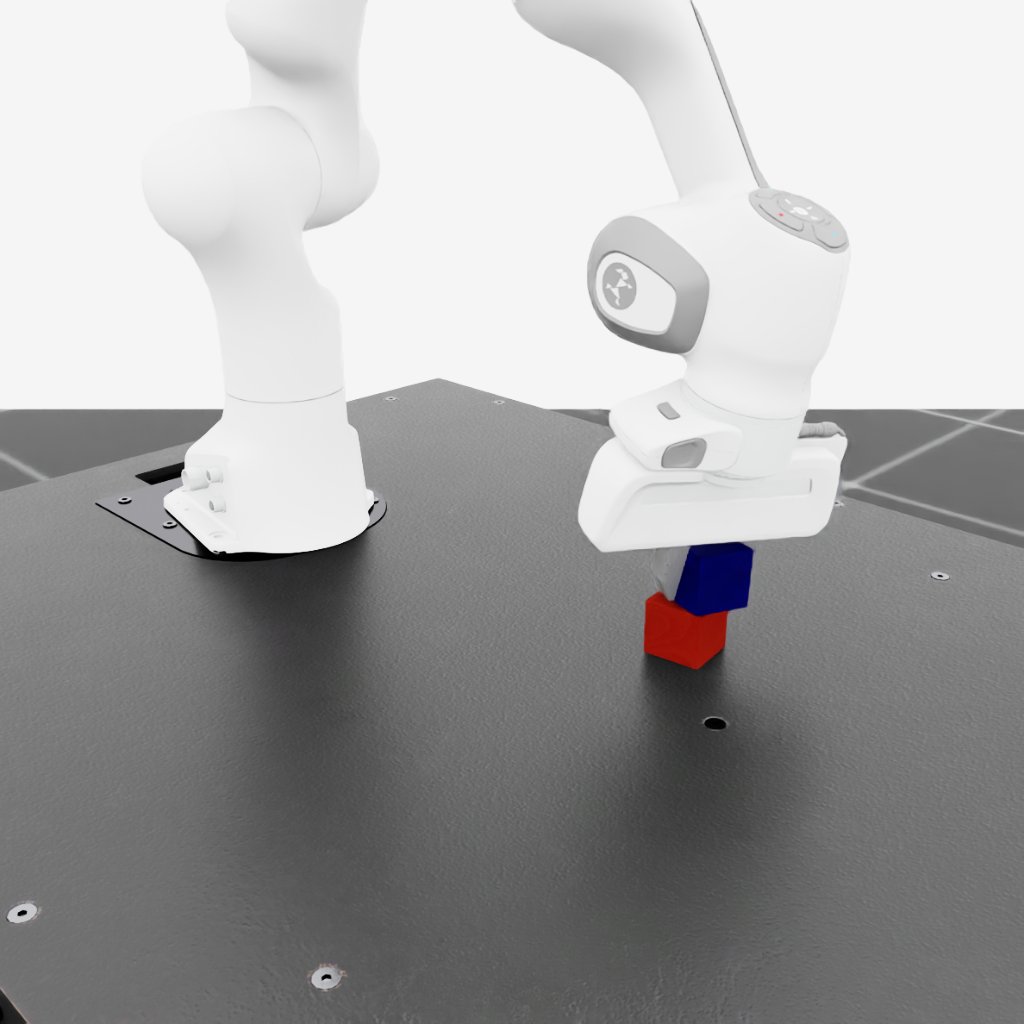}{0.33} &
      \imwjpg{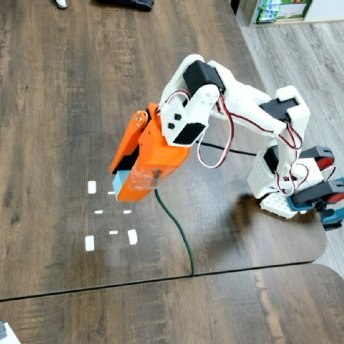}{0.33} \\
      (c) Nut threading & (d) Box placement & (f) Cube pickup
    }
  \end{adjustbox}
  \caption{
    \textbf{Task suites in simulation and the real world.} Four contact-rich manipulation tasks in simulation (a--d) and two real-world tasks (e, f) that mirror the simulated peg insertion and cube pickup.
  }
  \label{fig:task_suite}
  \vspace{-5pt}
\end{figure}

\section{Experiments}
\label{sec:experiment}

We evaluate \method{} on downstream RL across four contact-rich manipulation tasks in simulation and in the real world. Our experiments are designed to answer the following questions: (1) Does \method{} improve the performance of downstream RL? (2) How does \method{} compare with visual reward model? (3) Does \method{} generalize to novel object instances? (4) Is a learned tactile reward necessary? (5) Does \method{} enable real-world RL?

\paragraph{Experiment Setup}
For each task, we train \method{} on successful and failed demonstrations collected at \textbf{a single object position} using RL in simulation or teleoperation in the real world. We evaluate the trained reward model along two axes: (1) \emph{reward quality}, the discrepancy between predicted rewards and ground-truth task progress on held-out trajectories; and (2) \emph{downstream RL efficiency}, the success rate of policies trained with \method{} as a shaping reward within a fixed budget of environment interactions, averaged over 3 seeds.  Because downstream policies are trained on a \textbf{disjoint set of object positions} from the one used to collect demonstrations, any benefit from \method{} reflects generalization to out-of-distribution object configurations. 

In simulation, we use a Franka Panda Robot and simulate tactile sensors with TacSL~\cite{akinola2025tacsl}.  In the real world, we use a LeRobot SO101~\cite{Knight_Standard_Open_SO-100}, equipped with PaXini PX-6AX-GEN1 tactile sensors.

\paragraph{Simulation task suite}
We evaluate \method{} on four manipulation tasks (\cref{fig:task_suite}). \textit{Box placement} is a pick-and-place task requiring a stable grasp and controlled release, it serves as our simplest contact setting. The remaining three tasks come from the Isaac Lab Factory and FORGE suites~\cite{mittal2025isaaclab}, spanning a spectrum of contact complexity: \textit{peg insertion} demands precise alignment under tight clearance; \textit{gear assembly} requires rotational meshing with sustained multi-point contact; and \textit{nut threading} involves helical contact trajectories with continuous force feedback.  The benchmark originally initializes each episode with the target object pre-grasped.  This oversimplifies contact-rich manipulation: grasping objects is challenging and manipulation tasks can easily fail without stable grasps.  Therefore, we initialize the scene with objects placed on the table, and train policies to perform a sequence of tasks: first reach, next grasp, and then insert, thread, or place.  Our setup tests \method{}'s ability to provide useful rewards across diverse contact phases.

% \begin{figure}[t]
%     \vspace{-18pt}
%     \centering
%     \begin{adjustbox}{width=0.5\linewidth}
%     \tb{@{}cc@{}}{1.0}{
%     \imw{figures/TaRL/box_TaRL}{0.5} &
%     \imw{figures/TaRL/peg_TaRL}{0.5} \\ [-2pt]
%     Box placement & Peg insertion \\
%     \imw{figures/TaRL/gear_TaRL}{0.5} & 
%     \imw{figures/TaRL/nut_TaRL}{0.5} \\ [-2pt]
%     Gear assembly & Nut threading
%     }
%     \end{adjustbox}
%     % \vspace{-6pt}
%     \caption{
%       \textbf{Downstream RL efficiency.}
%       We evaluate RL policies trained \textcolor{blue}{with} and \textcolor{gray}{without} \method{} in simulation.  \method{} substantially enhances the training efficiency and task success.
%     }
%     \label{fig:TaRL}
%     % \vspace{-12pt}
% \end{figure}

\begin{figure}[]
    \centering
    \footnotesize
    % \vspace{-18pt}
    \begin{adjustbox}{width=\linewidth}
    \tb{@{}cccc@{}}{0.1}{
    \imw{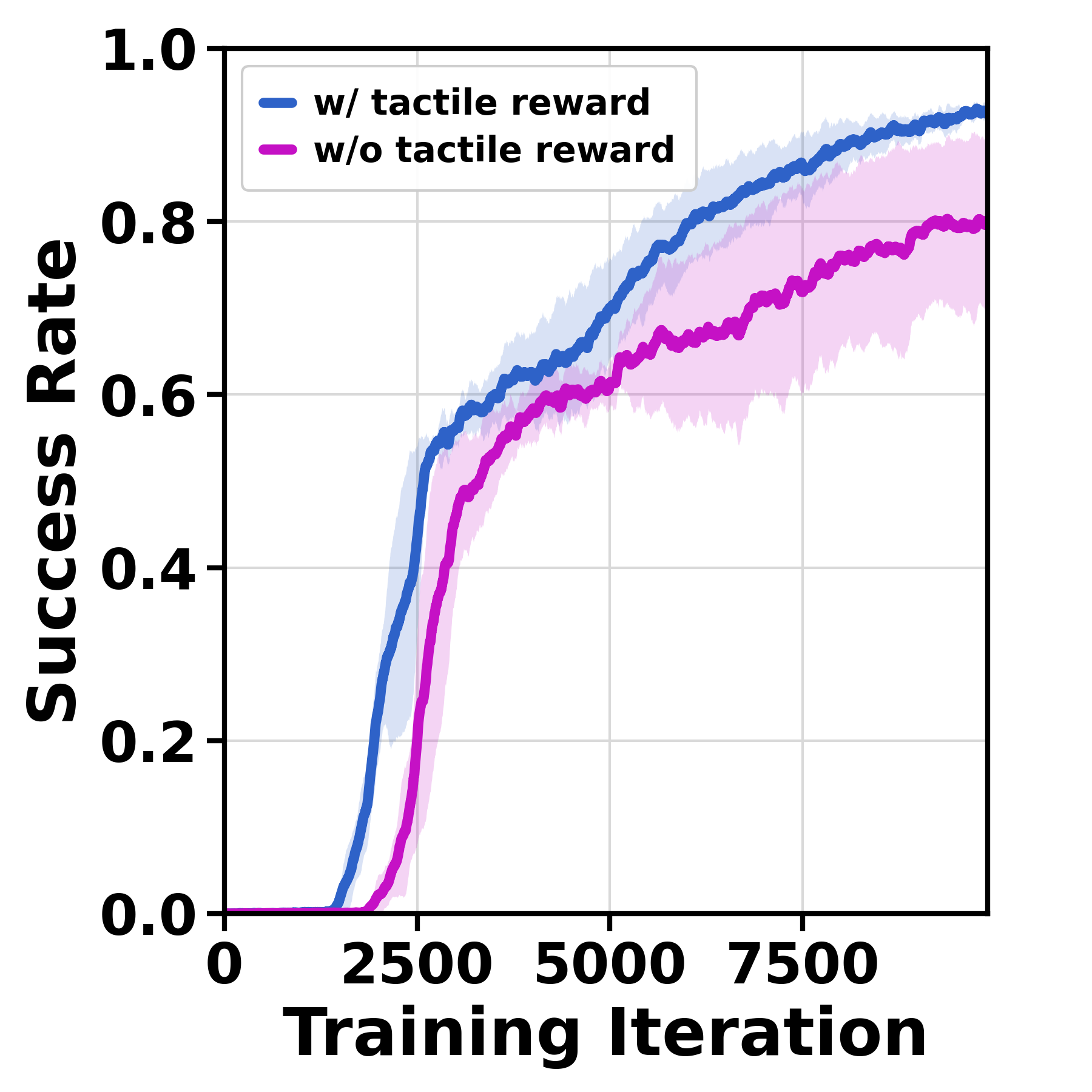}{0.25} &
    \imw{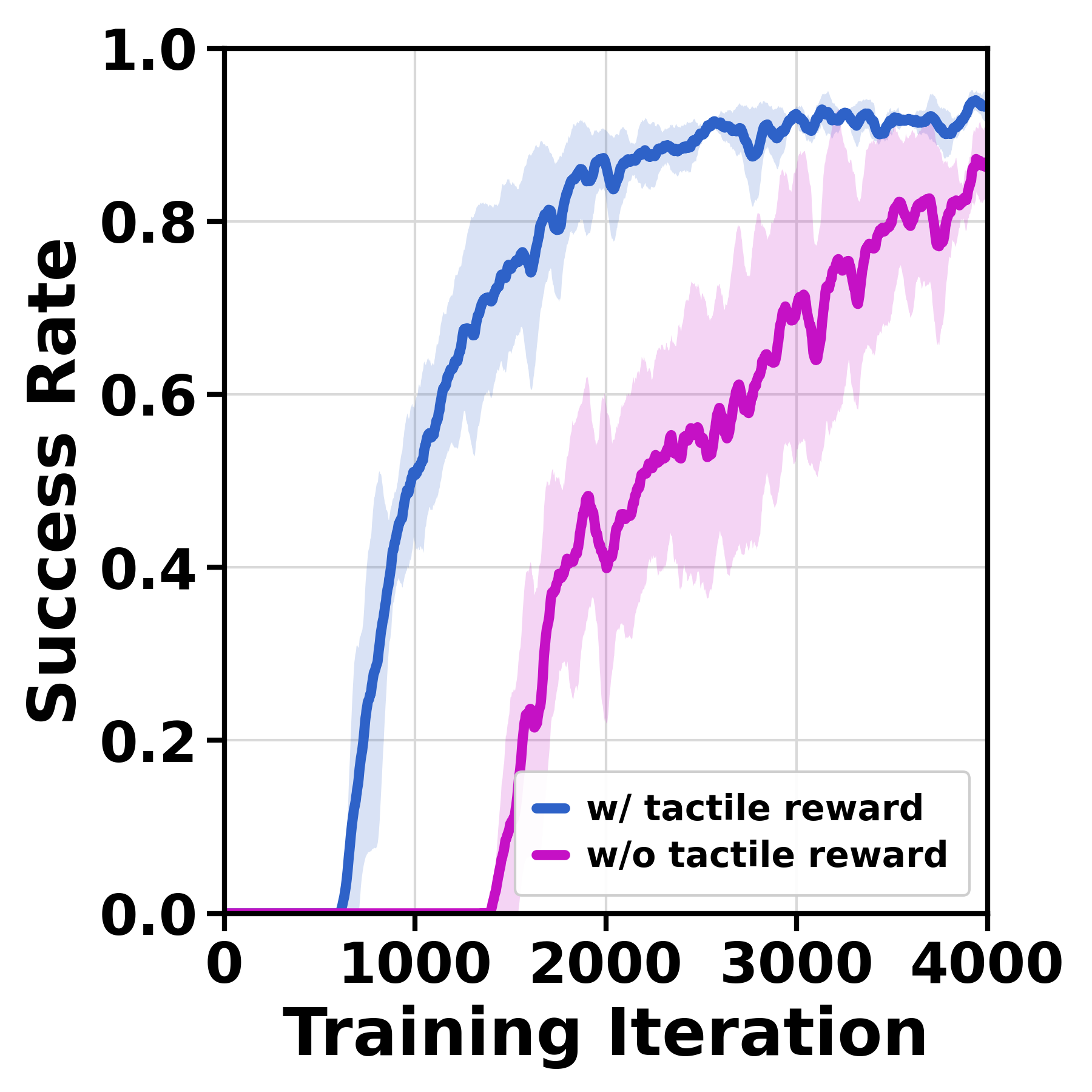}{0.25} &
    \imw{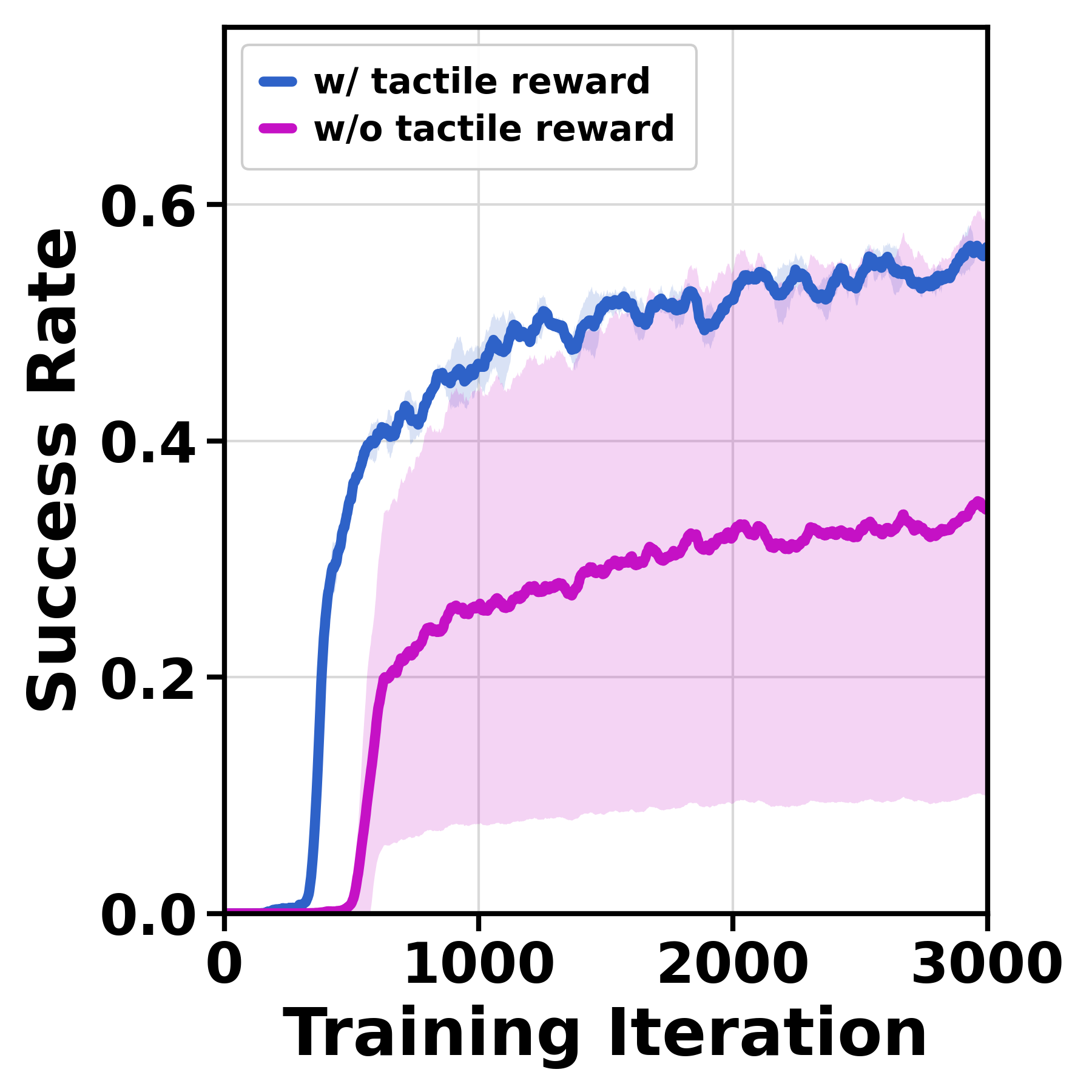}{0.25} & 
    \imw{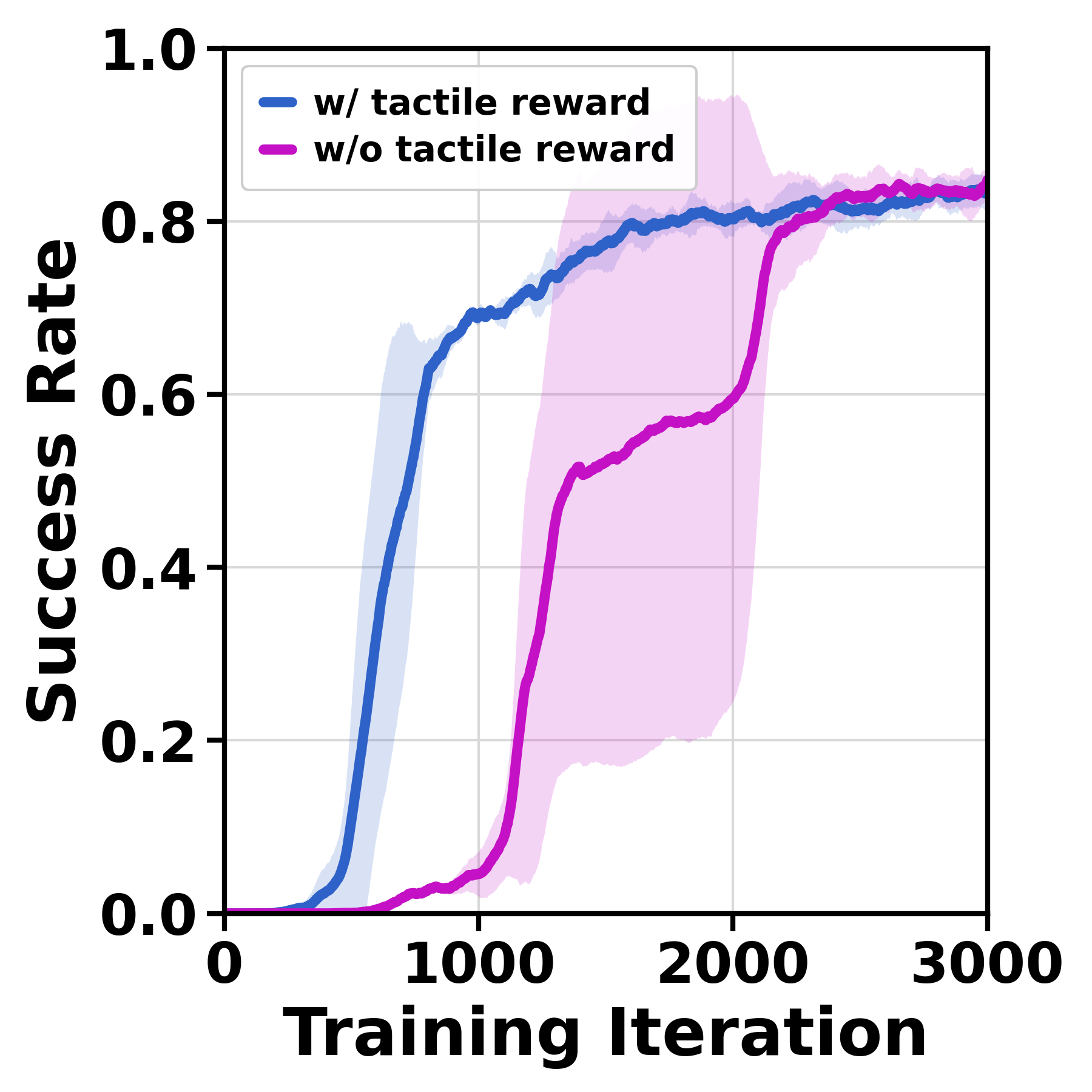}{0.25} \\
    Box placement & Peg insertion &
    Gear assembly & Nut threading
    }
    \end{adjustbox}
    \caption{
      \textbf{Downstream RL efficiency.}
      We evaluate RL policies trained \textcolor{tarlblue}{with} and \textcolor{tarlpink}{without} \method{} in simulation.  \method{} substantially enhances the training efficiency and task success.
    }
    \label{fig:TaRL}
    \vspace{-12pt}

\end{figure}

\subsection{Does \method{} improve downstream RL?}
\label{subsec:TaRL}
We validate whether \method{} enhances downstream RL with the four manipulation tasks in simulation.  For each task, we train \method{} with 200 successful and 800 failed demonstrations, and test its downstream RL efficiency and reward quality.

We first compare RL training efficiency with and without \method{} as a shaping reward.  We devise simulation environments providing basic reward functions for reaching, lifting, transporting, and placing the target object, but deliberately excludes terms encouraging stable grasp or precise contact---the capability we expect \method{} to enable.  Visual rewards are ignored in this experiment ($\beta =0$).  For box placement, peg insertion, gear assembly and nut threading, we set tactile reward weightings $\alpha$ to $0.2$, $0.175$, $0.175$ and $0.175$, respectively.  All results are summarized in \cref{fig:TaRL}.  \method{} substantially enhances RL's learning efficiency and the final success rate for most tasks.  Especially on nut threading, the task success increases from 34\% to 56\% using \method{}.

\begin{figure}[t]
    \centering
    \footnotesize
    % \vspace{-18pt}
    \begin{adjustbox}{width=\linewidth}
    \tb{@{}cccc@{}}{0.1}{
    \imw{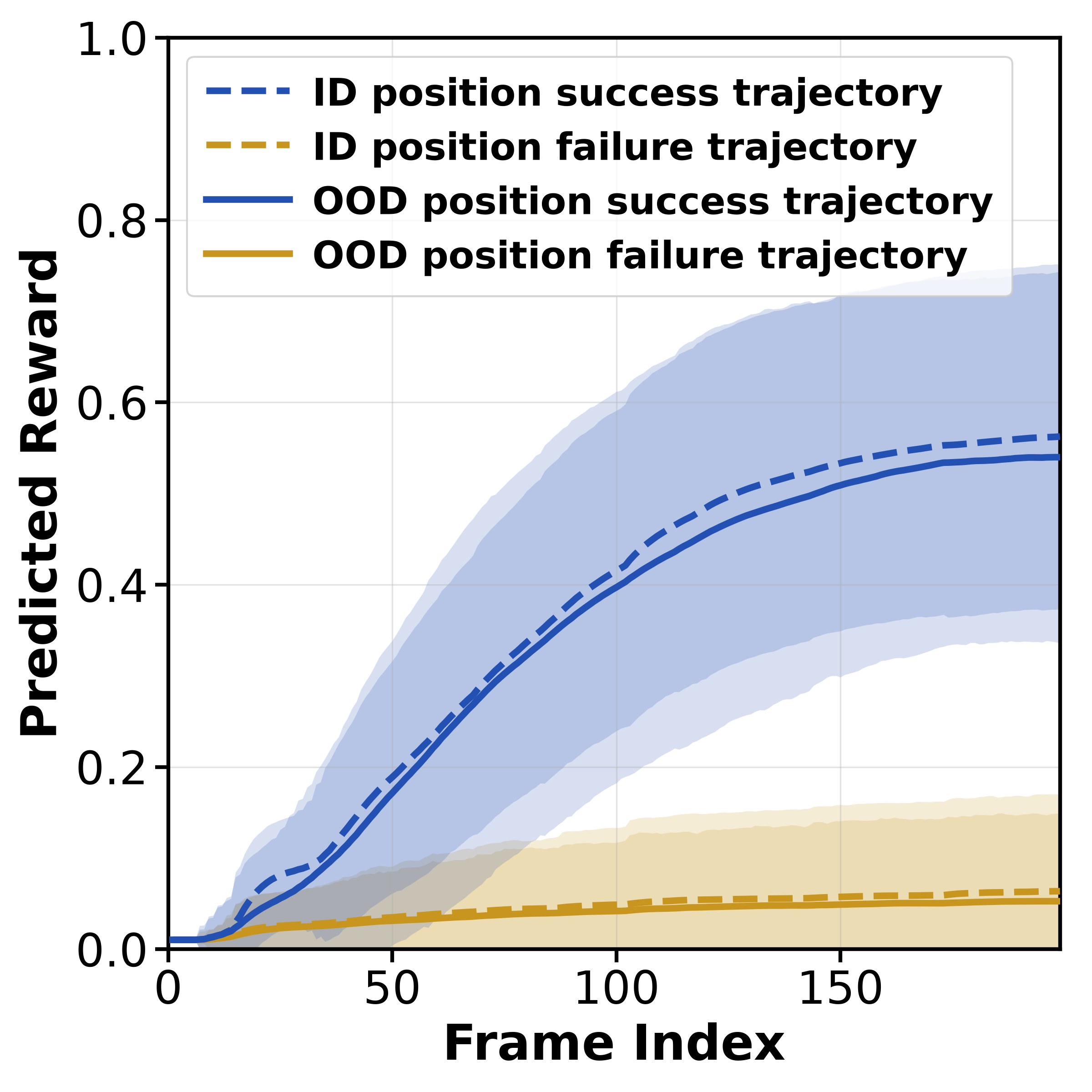}{0.25} &
    \imw{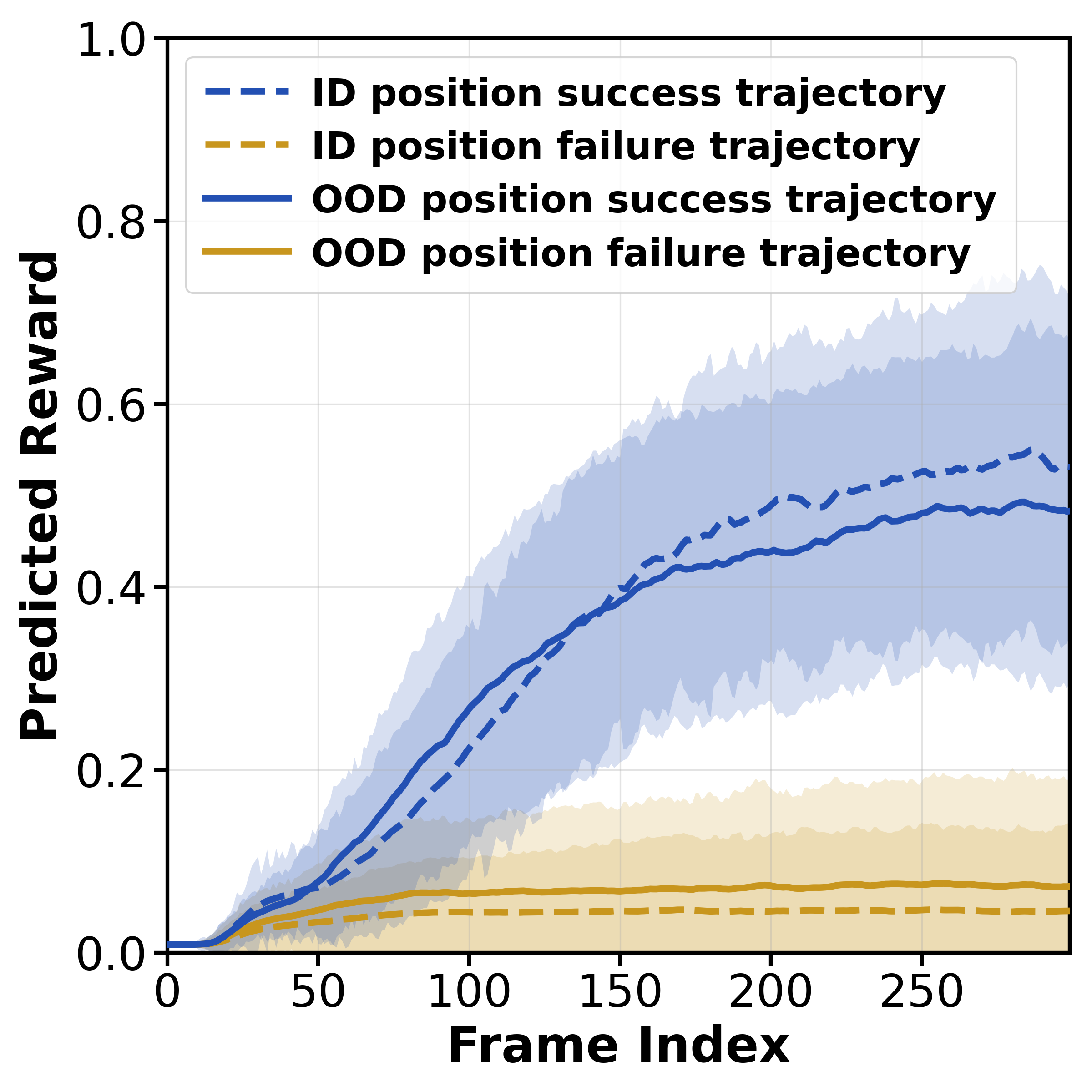}{0.25} &
    \imw{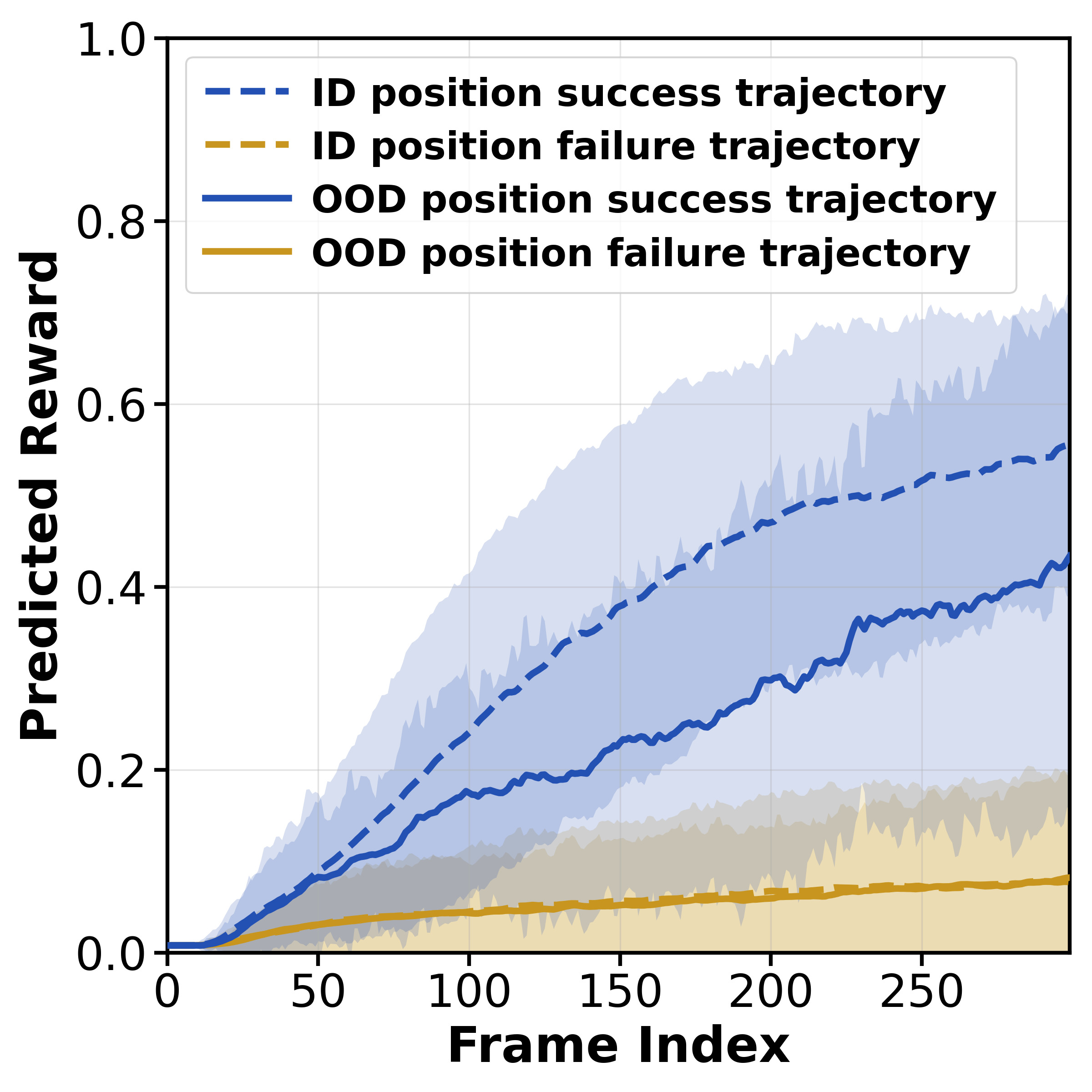}{0.25} & 
    \imw{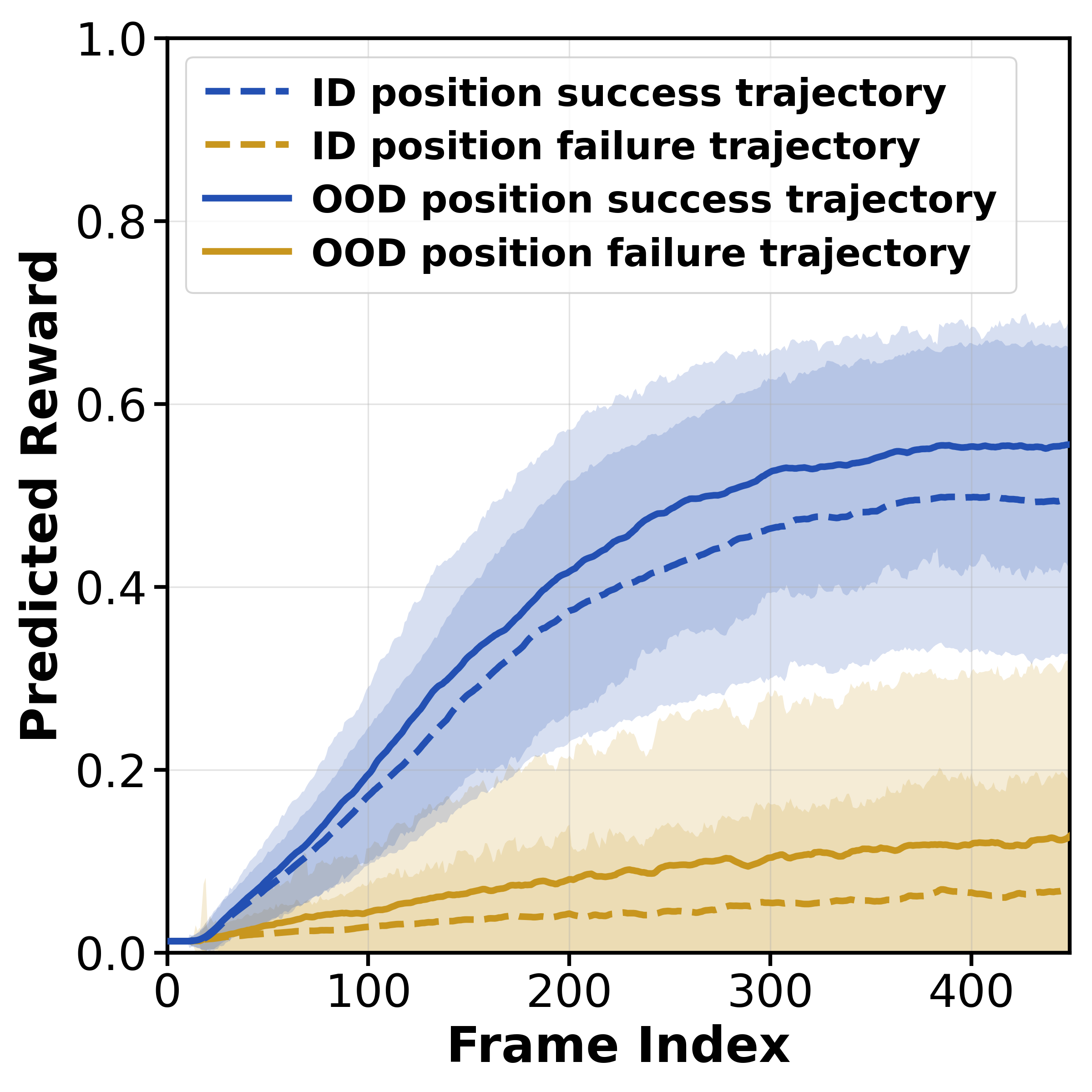}{0.25} \\
    Box placement & Peg insertion &
    Gear assembly & Nut threading
    }
    \end{adjustbox}
    \caption{
      \textbf{Generalization to unseen object positions.}  We train \method{} on a single position, and deploy it to seen (ID) and unseen (OOD) object positions.  We report the predicted reward at each time step, averaged within each configuration.  \method{} is robust to OOD object positions.
    }
    \label{fig:rolling}
    \vspace{-12pt}

\end{figure}

We next evaluate \method{}'s reward quality.  The model is trained on demonstrations collected at a single object position, and evaluated along two axes:  (1) \textbf{In-domain (ID)} generalization tests \method{} on held-out trajectories collected with the same object configurations as training;  and (2) \textbf{Out-of-distribution (OOD)} generalization tests \method{} on trajectories collected at novel object configurations.  For both cases, we split the test set into a set of 50 successful trajectories and a set of 150 failed trajectories; meanwhile, we report the predicted reward at each time step, averaged within each set.  A larger gap between the predicted rewards across these two sets means that the learned reward function can more easily distinguish between successful and failed trajectories, thereby allowing RL policies to pick out high-value samples from collected rollouts during training.  As shown in \cref{fig:rolling}, \method{} predicts much higher rewards on successful trajectories than failed ones.  Moreover, \method{} demonstrates strong generalization to OOD object configurations.

\begin{figure*}[t]
    \centering
    \vspace{-10pt}
    \begin{adjustbox}{width=\linewidth}
    \tb{@{}ccc@{}}{0.1}{
    \imw{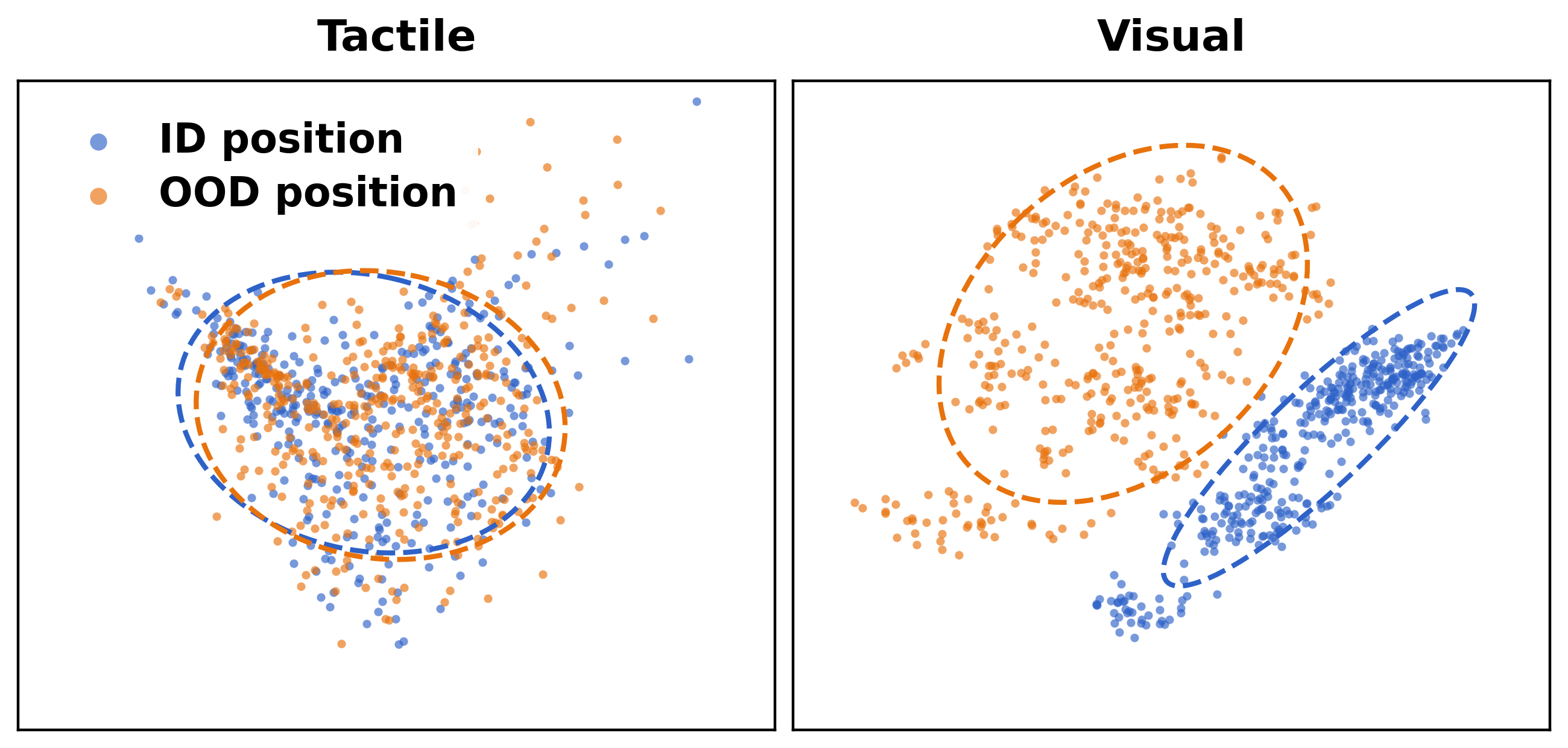}{0.33} & \imw{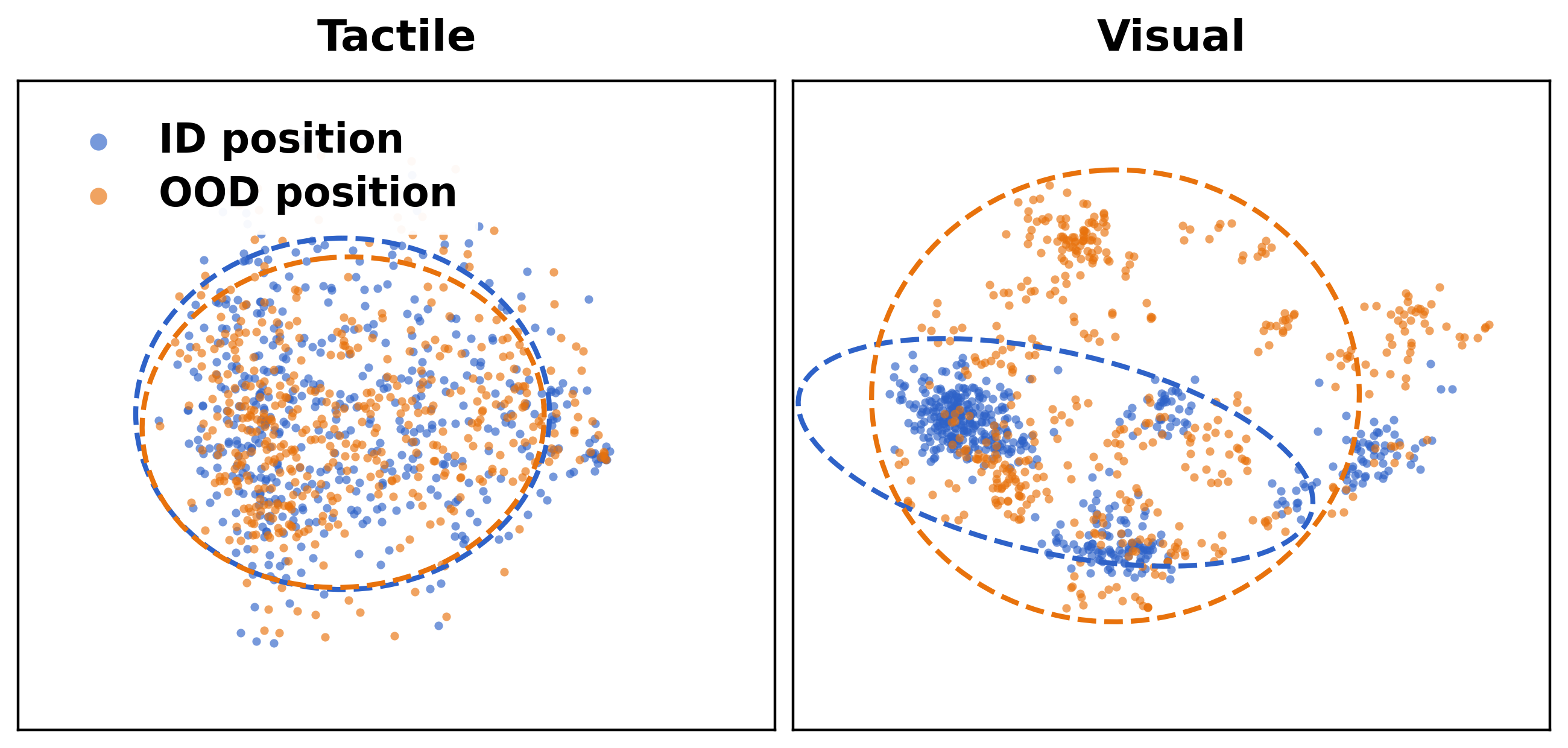}{0.33} & \imw{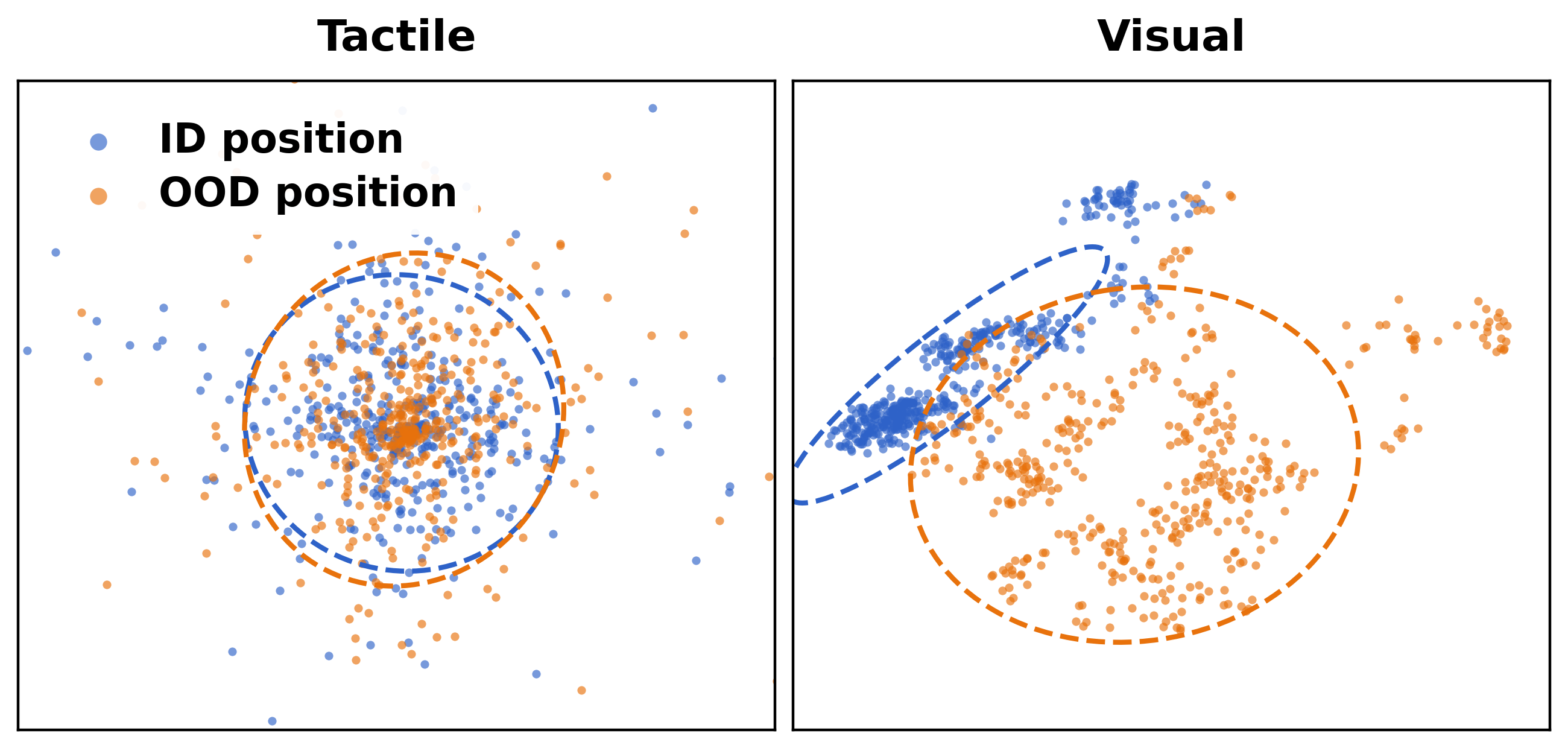}{0.33} \\ [-2pt]
    Peg insertion &
    Gear assembly & Nut threading
    }
    \end{adjustbox}
    \begin{minipage}{\textwidth}
    \captionsetup{width=\textwidth,margin=0pt}
    \caption{
      \textbf{Comparison of tactile and visual features.}  We extract features with the CNN of ReWiND and \method{}, each taking visual and tactile observations as input, on peg insertion, gear assembly and nut threading tasks.  With t-SNE, features are transformed into 2-dim vectors and plotted in 2D.  Dashed ellipses denote covariances of their distributions.  Visual features extracted from \textcolor{blue}{in-domain (ID)} object positions follow distinct distributions from those of \textcolor{orange}{out-of-distribution (OOD)} positions.  In contrast, tactile features in both cases share similar distributions.
    }
    \label{fig:Tactile_Vis_Traj_Distribution}
    \end{minipage}
    \vspace{-15pt}

\end{figure*}

\subsection{How does \method{} compare with visual reward model?}
\label{subsec:visual}
To better understand tactile rewards, we systematically compare \method{} against a seminal visual reward learning framework---ReWiND~\cite{zhang2025rewind}.  Concretely, we evaluate their OOD generalization, analyze their output rewards contributed in individual contact phases, and study whether they provide complementary rewards which further enhance RL efficiency.  Experiments are conducted on peg insertion, gear assembly and nut threading tasks.

\begin{figure}[H]
    \centering
    \vspace{-5pt}
    \begin{adjustbox}{width=0.9\linewidth}
    \tb{@{}ccc@{}}{0.1}{
    \imw{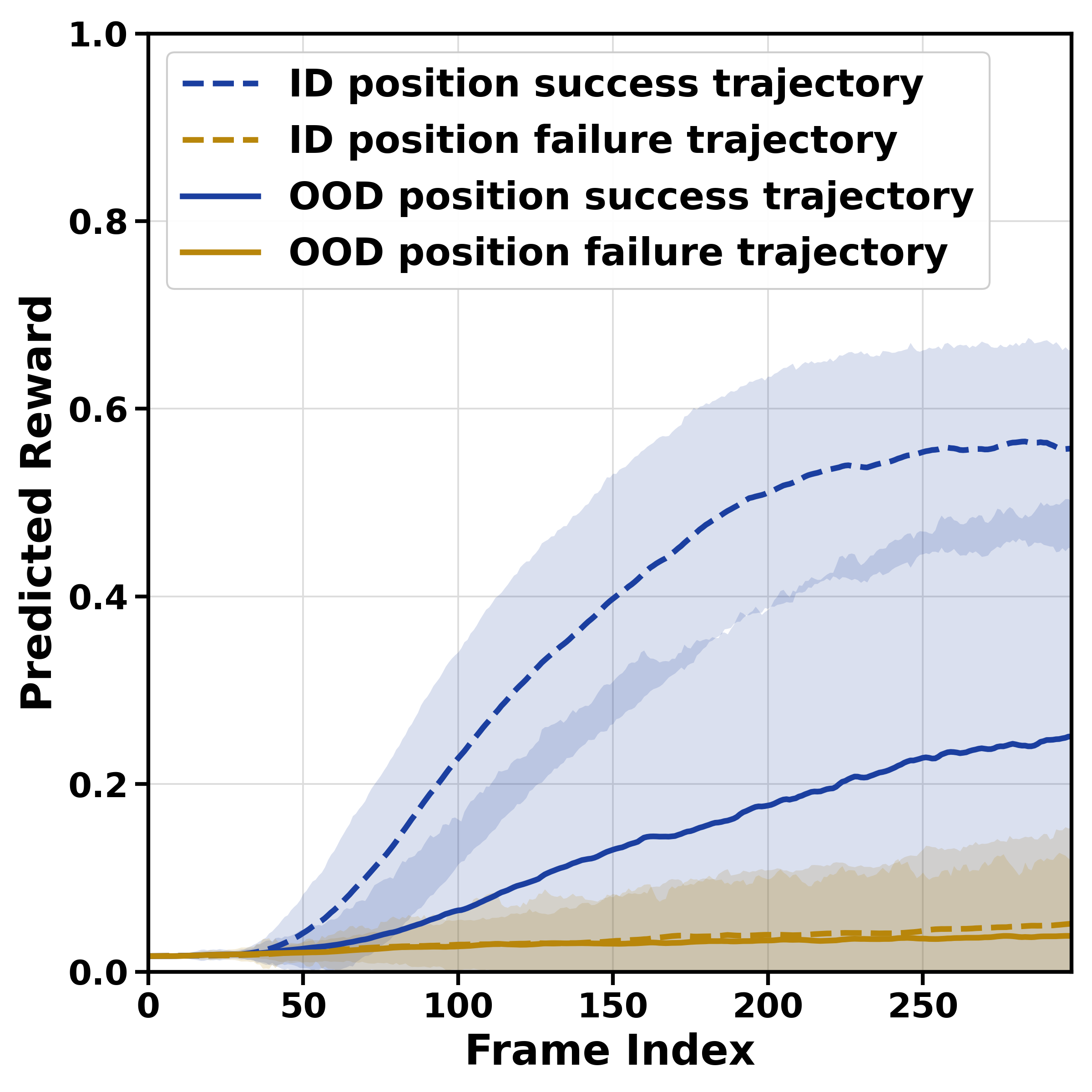}{0.33} &
    \imw{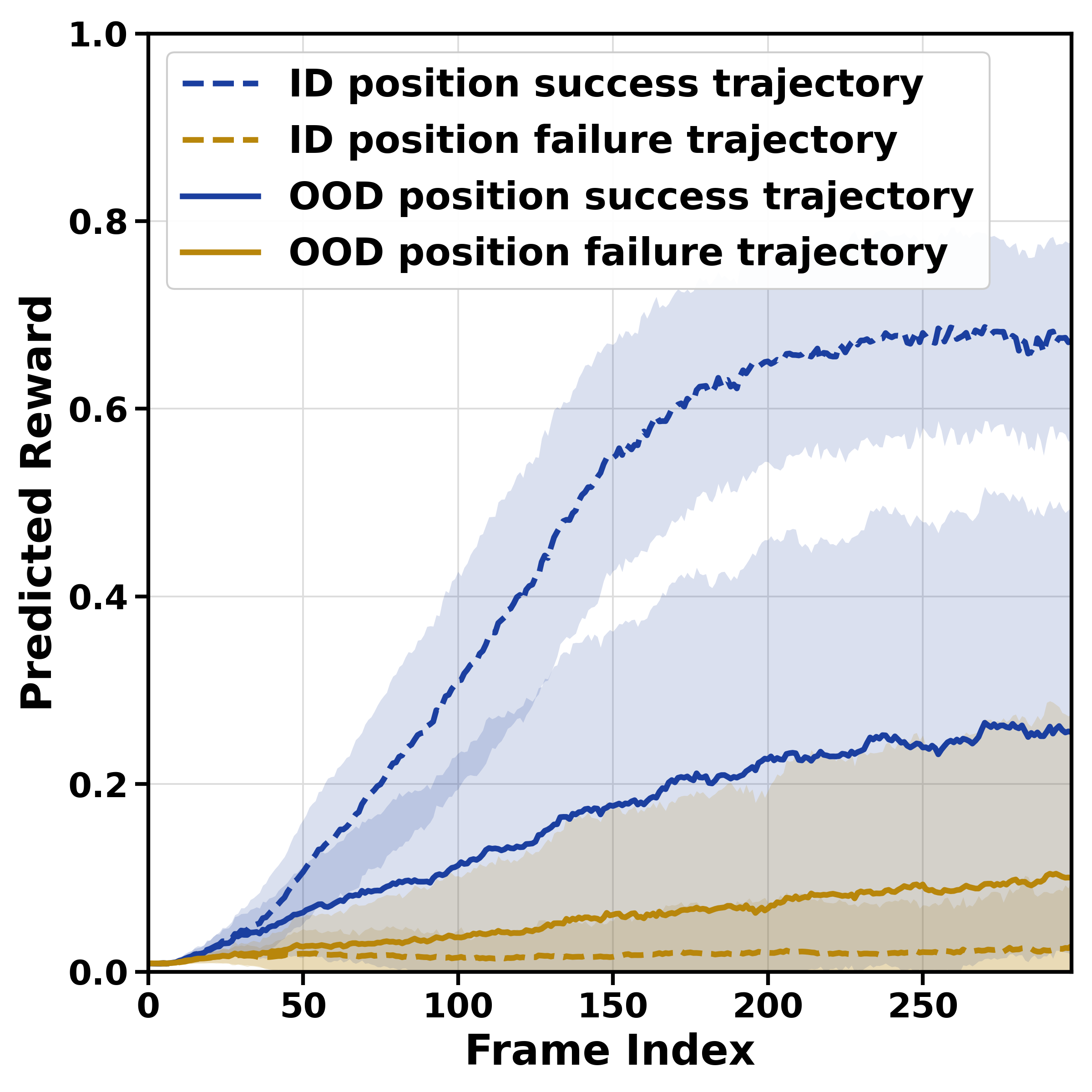}{0.33} & \imw{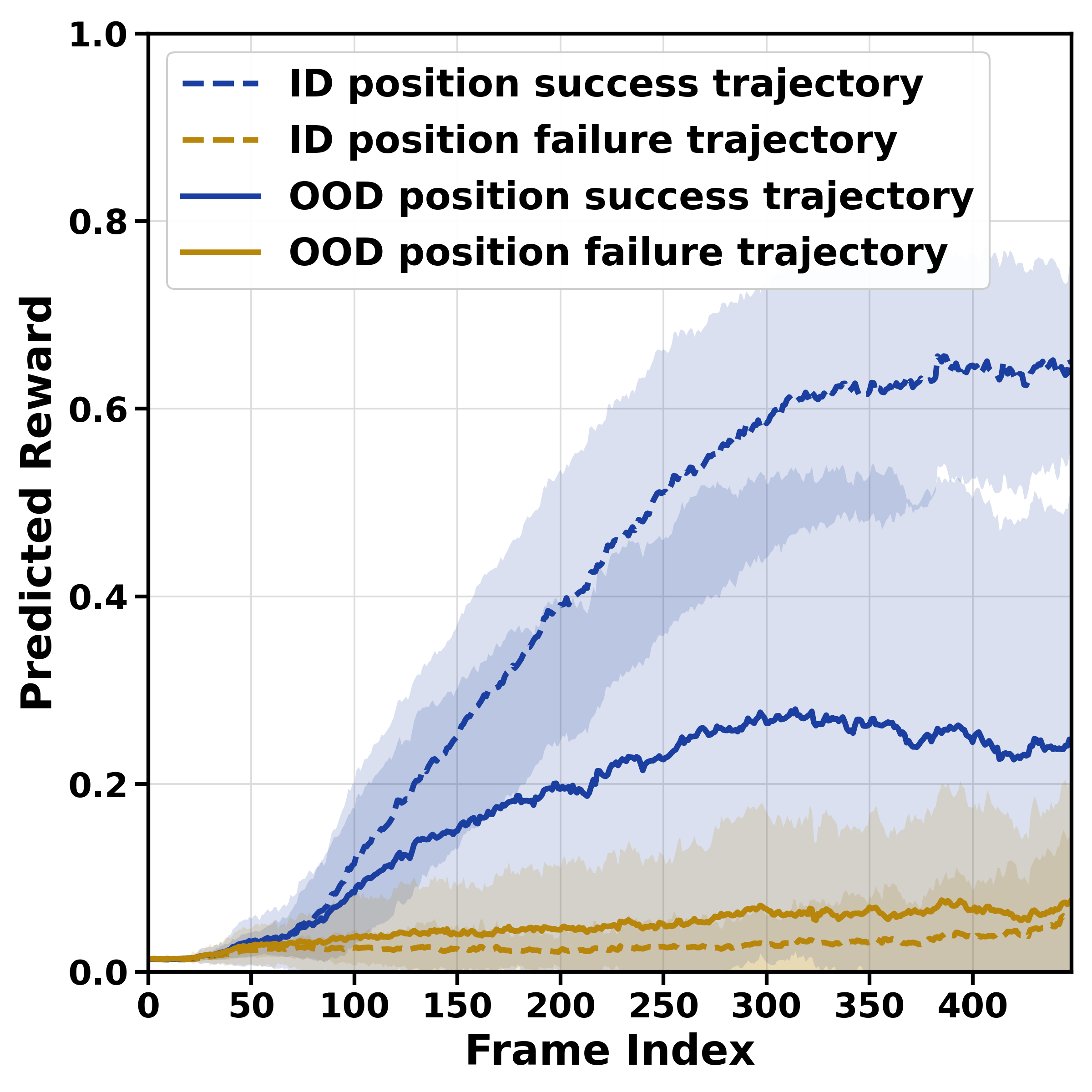}{0.33} \\ [-2pt]
    \imw{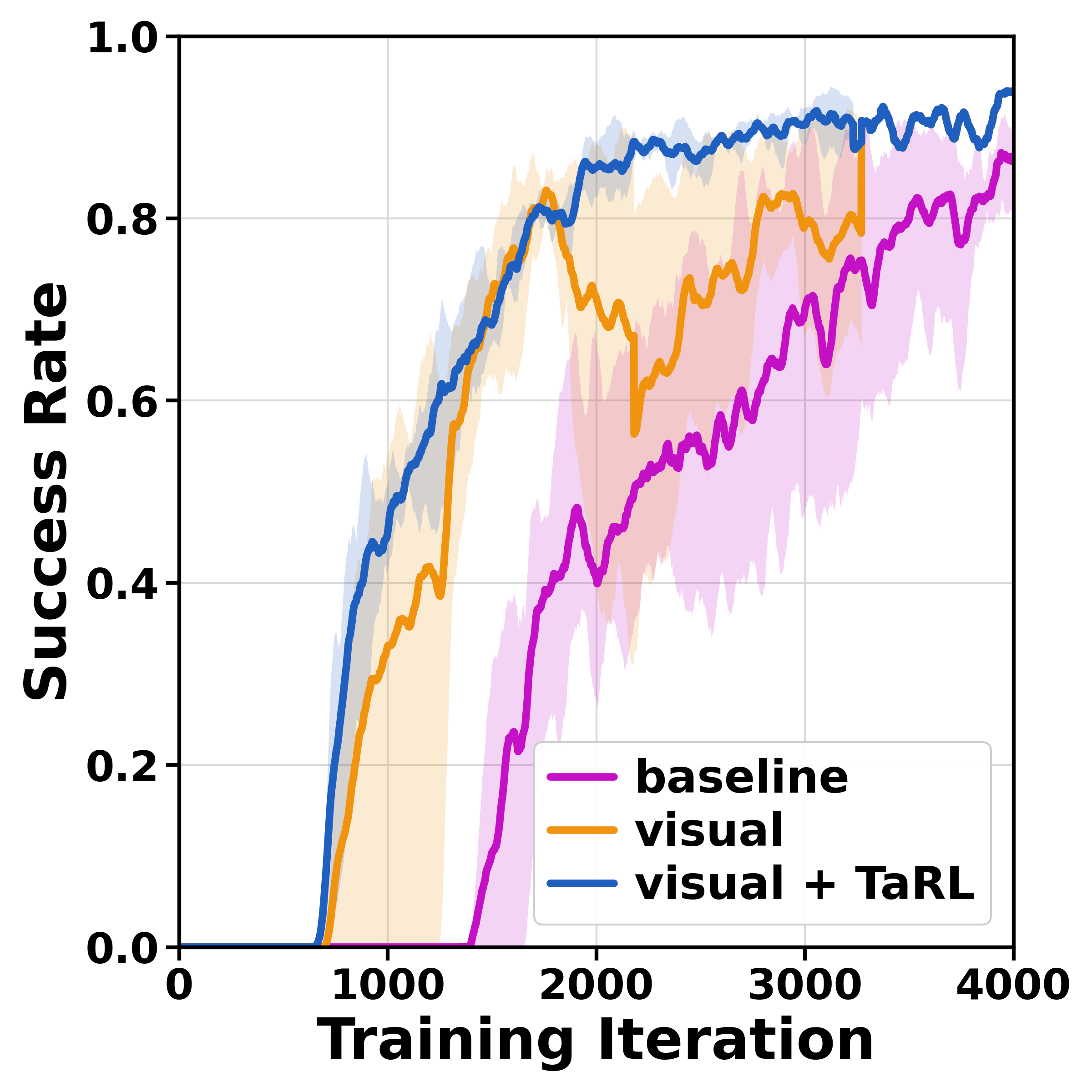}{0.33} &
    \imw{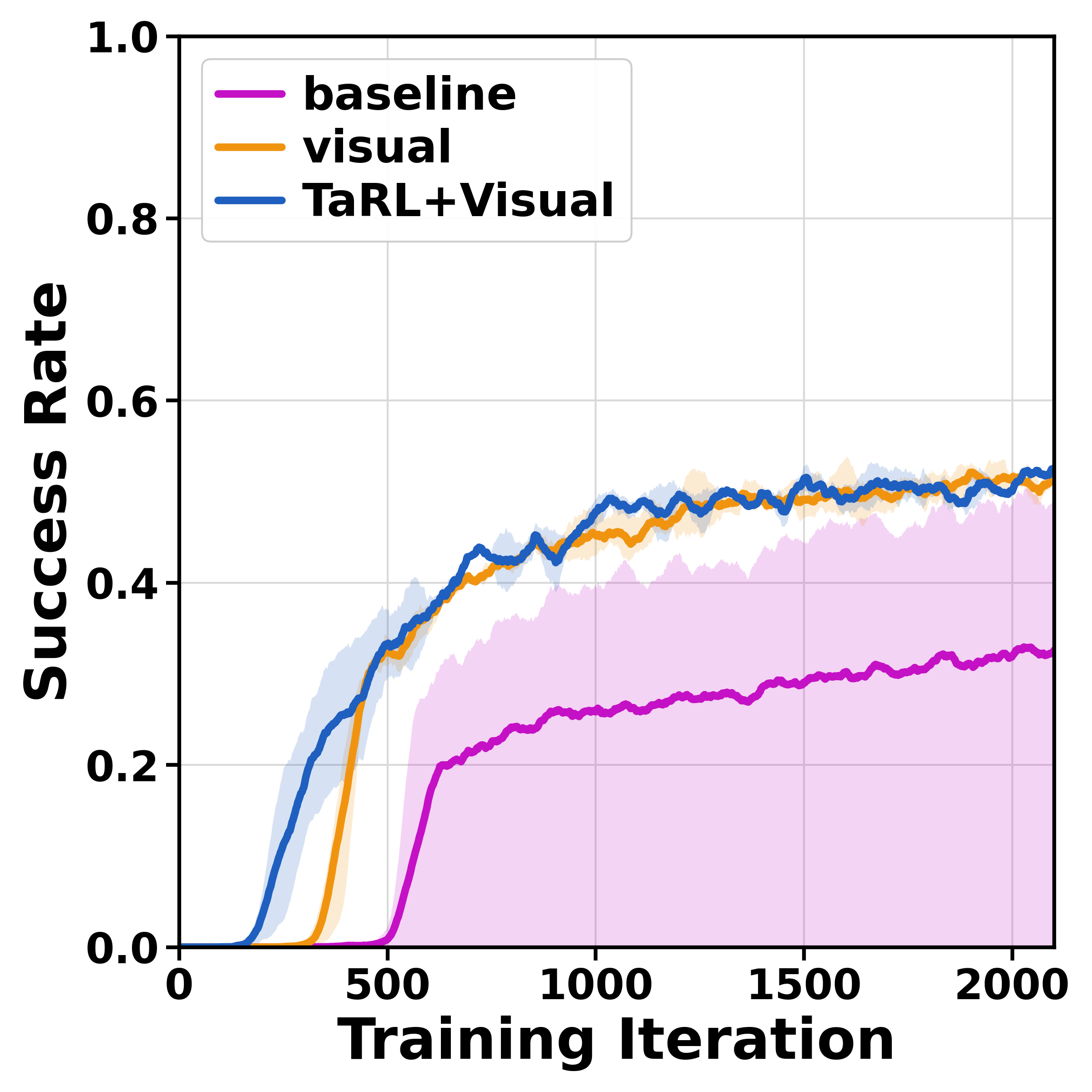}{0.33} & \imw{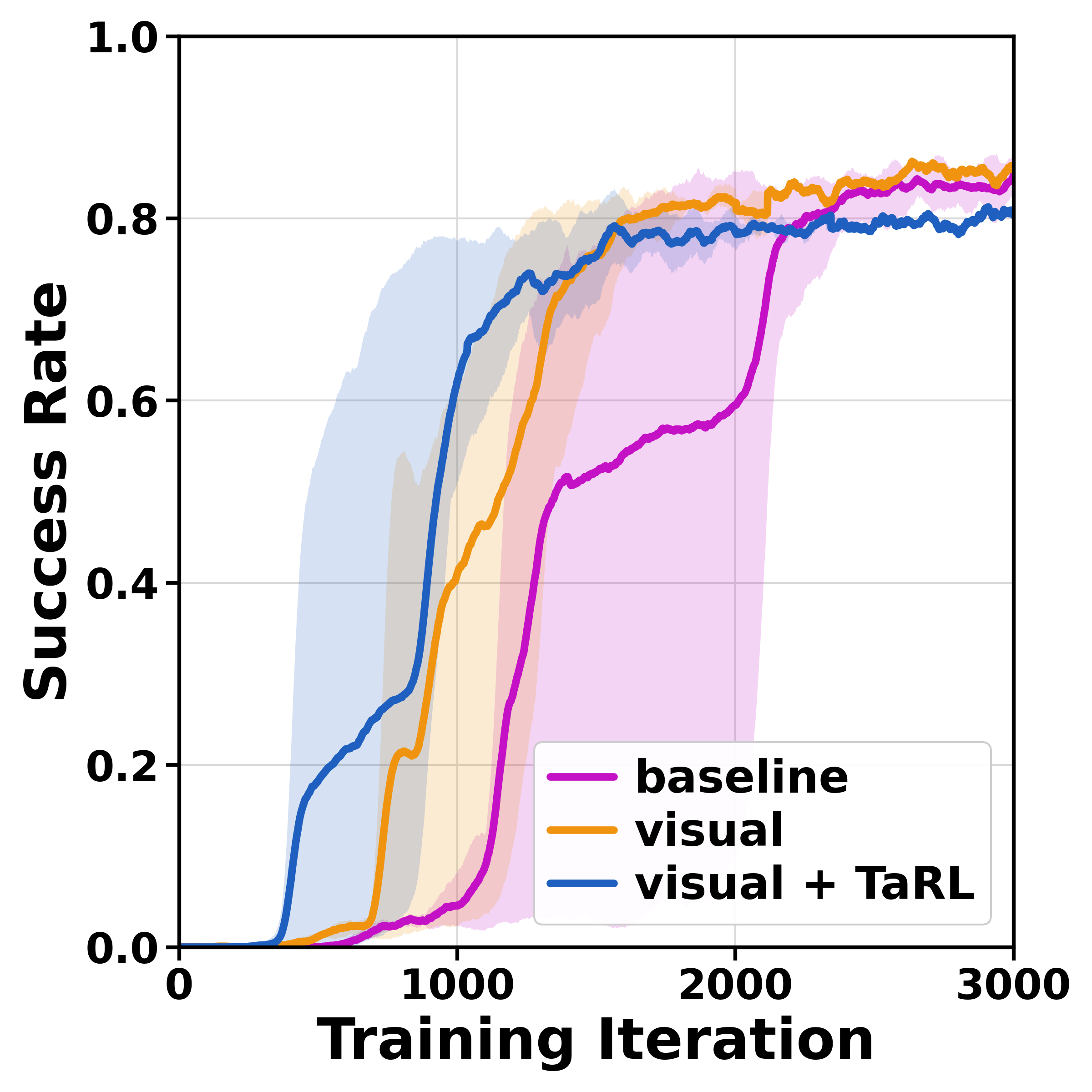}{0.33} \\ [-2pt]
    Peg insertion &
    Gear assembly & Nut threading
    }
    \end{adjustbox}
    \captionsetup{skip=3pt}
    \caption{
      \textbf{Evaluation of visual rewards.} \textbf{Top:} Visual reward quality on ID and OOD object positions. \textbf{Bottom:} Downstream RL efficiency of visual rewards and the combination of tactile and visual rewards.   
    }
    \label{fig:TaRL_Visual}
    \vspace{-8pt}
\end{figure}

\begin{figure*}[t]
    \vspace{-10pt}
    \centering
    \includegraphics[width=\linewidth]{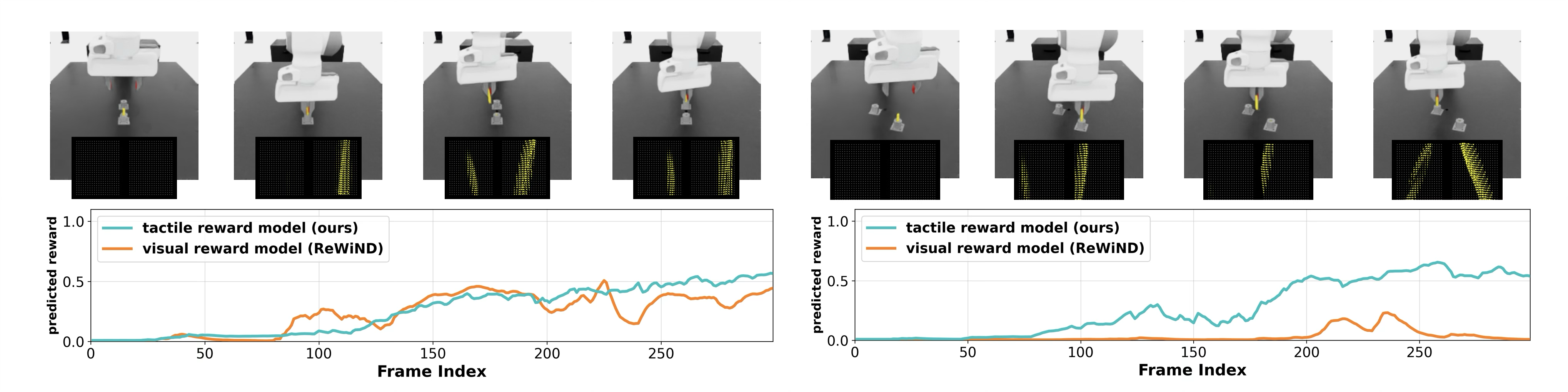}
    \caption{\textbf{The visual and tactile reward model's generalization to unseen object positions.}  We show predicted tactile and visual rewards along successful rollouts at the training position (left) and an unseen position (right). The single-position visual model collapses at the unseen position while TaRL does not.}
    % \caption{\textbf{Visual reward model trained at a single position.} Predicted tactile and visual rewards along successful rollouts at the training position (left) and an unseen position (right). The single-position visual model collapses at the unseen position while TaRL does not.}
    \label{fig:episode_visual_tactile_reward}
    \vspace{-10pt}
\end{figure*}

\paragraph{OOD generalization}
We train ReWiND under two setups: the first uses the same set of demonstrations as \method{} collected from a single object position, and the other separately collects demonstrations from multiple object positions.  We evaluate both models' reward quality given ID and OOD object configurations.  As shown in \cref{fig:episode_visual_tactile_reward}, while ReWiND trained with a single position can handle seen configurations, it fails to generalize to OOD positions.  Even trained with multiple object positions, the visual reward quality degrades substantially with OOD positions: the gap between the predicted rewards across successful and failed trajectories decreases from 0.5 to 0.1 after initializing the scene with unseen object configurations (the top of \cref{fig:TaRL_Visual}).

To understand why ReWiND fails to generalize, we further analyze feature tokens $d$ encoded from different input modalities.  Given a trajectory, we uniformly sample 16 intermediate frames and encode each frame into a latent token with the CNN of \method{} / ReWiND.  Then, we visualize these tokens by transforming them into 2-dim feature vectors with t-SNE~\cite{van2008visualizing} and plotting them in 2D.  As shown in \cref{fig:Tactile_Vis_Traj_Distribution}, tactile tokens extracted at OOD object positions share the similar distributions as those extracted at ID positions.  In contrast, visual tokens extracted at ID and OOD positions follow distinct distributions.  These results demonstrate that tactile data captures local robot-object interaction, and it is therefore robust to changes in scene layouts.

Because ReWiND trained on a single object position fails to produce robust rewards, we use the model trained on multiple positions for the following experiments.

\begin{figure}[H]
    \centering
    \includegraphics[width=\linewidth]{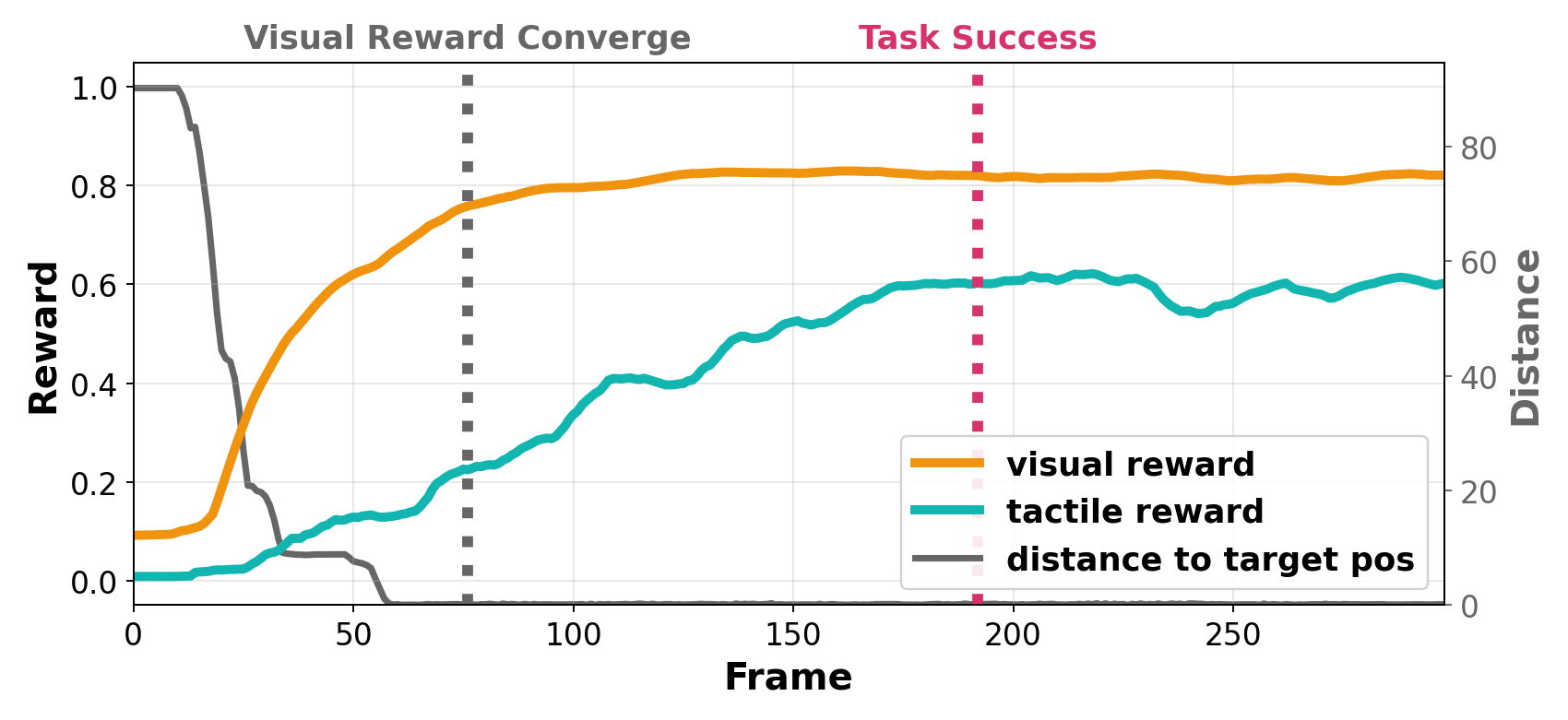}
    \caption{
      \textbf{Rewards contributed at individual contact phases.}  We show the \textcolor{orange}{visual} and \textcolor{cyan}{tactile} rewards predicted by ReWiND and \method{} on a trajectory of gear assembly task.  Visual rewards plateau once the robot reach the target object's position.  Tactile rewards vary across different contact phases.      
    }
    \label{fig:episode_reward_visualize}
    \vspace{-5pt}
\end{figure}

\paragraph{Rewards contributed in individual contact phases}
We plot visual and tactile rewards at each time step of a trajectory to analyze their contributions at individual manipulation stages. As shown in \cref{fig:episode_reward_visualize}, with a successful trajectory from \textit{gear assembly} task, the visual reward rises rapidly and halts as soon as the robot's end-effector reaches the target object's position.  This shows that the visual reward emphasizes the spatial relationship of the robot and the scene: it is useful to locate objects or decide moving directions, but the quality degrades at contact when local robot-object interaction cannot be observed visually.  In contrast, the tactile reward grows smoothly until the robot achieves task success.  It provides informative feedback at distinct contact phases.

\paragraph{Complementary reward shaping}
We study whether visual rewards complement tactile rewards, and whether combining both enhances RL efficiency.  We conduct experiments on peg insertion, gear assembly and nut threading tasks, setting both $\alpha$ and $\beta$ to 0.175.  We evaluate downstream RL efficiency, and present all results in the bottom of \cref{fig:TaRL_Visual}.  We show that combining tactile rewards with visual rewards accelerates downstream RL training.  For peg insertion task, they increase the task success substantially.

\begin{figure}[H]
    \centering
    \vspace{-10pt}
    \begin{adjustbox}{width=0.7\linewidth}
    \tb{@{}cc@{}}{0.1}{
    \imw{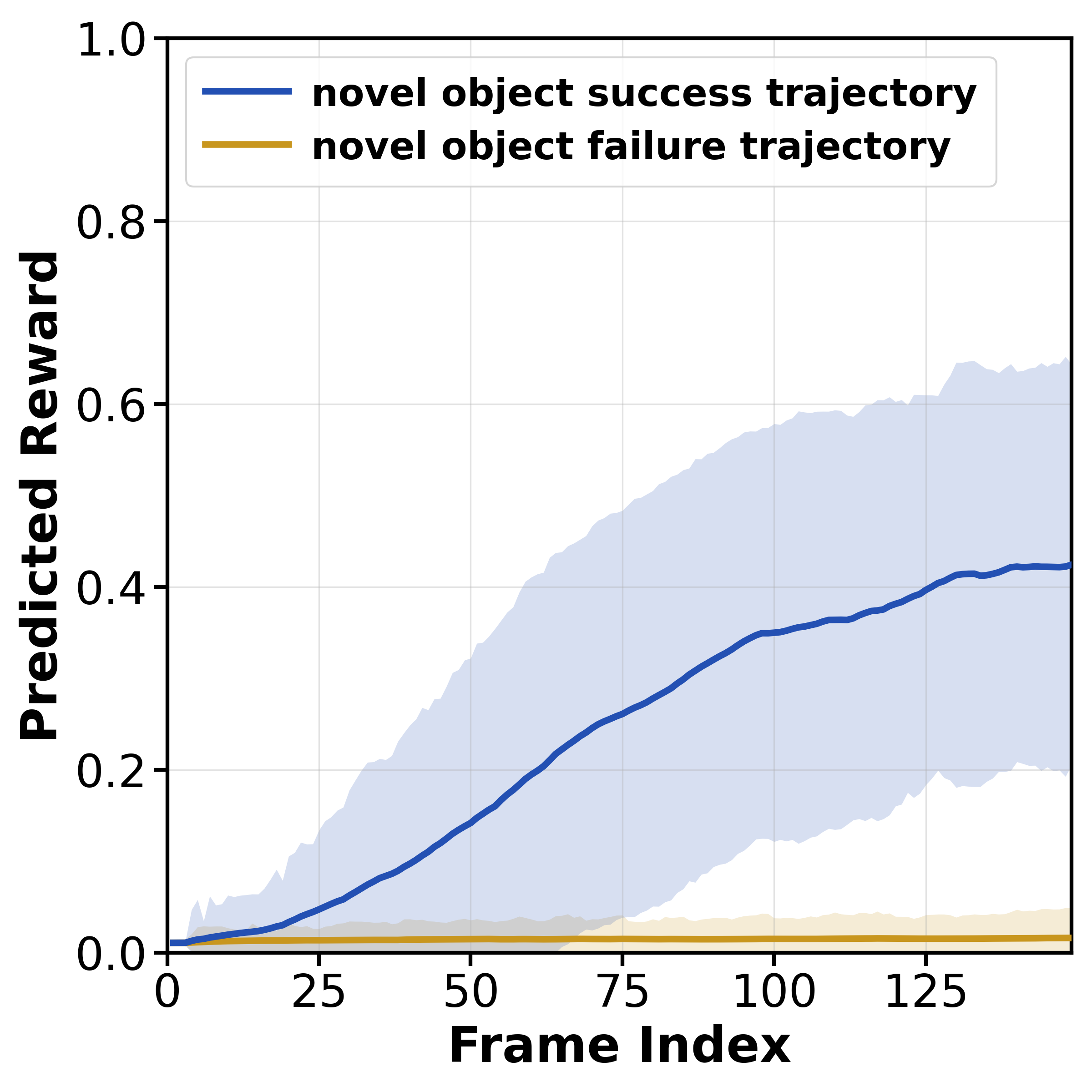}{0.45} &
    \imw{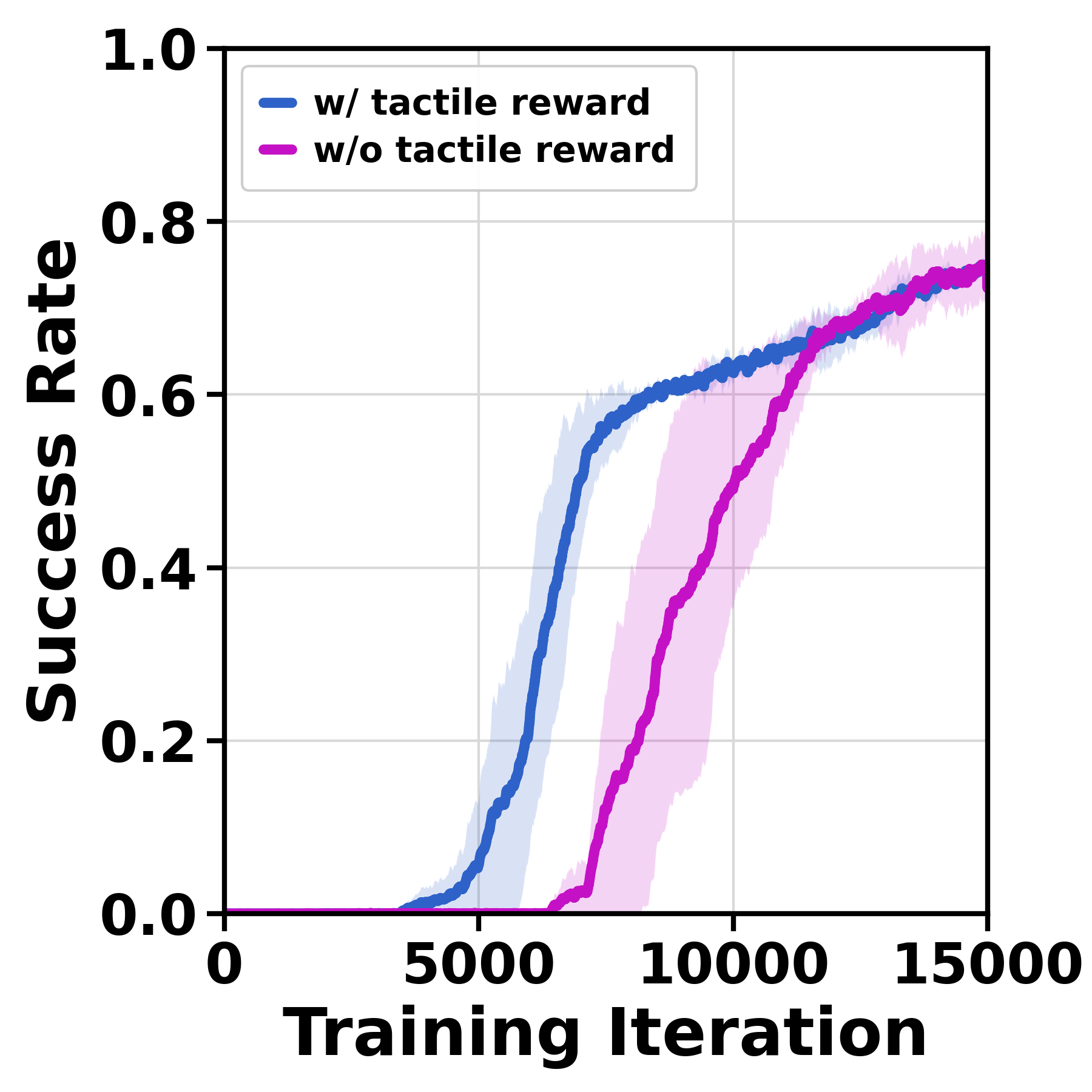}{0.45} \\ [-2pt]
    Reward Quality & Downstream RL 
    }
    \end{adjustbox}
    \caption{
      \textbf{Zero-shot generalization to unseen object instances.}  We deploy \method{} trained on box placement to can placement.  \method{} achieves high reward quality and improves downstream RL efficiency on the new task.}
    \label{fig:task_generalization}
    \vspace{-20pt}
\end{figure}

\subsection{Does \method{} generalize to novel object instances?}
\label{subsec:novel_object}
Beyond OOD object positions, we study whether tactile rewards capturing local interaction also generalize to OOD object instances.  We deploy \method{} trained on box placement task to \textit{can placement}, a task for picking and placing a can, without extra training.  We evaluate the model's reward quality and downstream RL efficiency on the unseen task during training. The left of \cref{fig:task_generalization} shows that the predicted rewards distinguish successful and failed trajectories; meanwhile, the right of \cref{fig:task_generalization} shows that the reward learning model zero-shot transfers to an unseen task, increasing the downstream RL efficiency significantly.

\subsection{Is a learned tactile reward necessary?}
\label{subsec:tactile_input}

\begin{figure}[t]
    \centering
    \vspace{-5pt}
    \begin{adjustbox}{width=\linewidth}
    \tb{@{}ccc@{}}{0.1}{
    \imw{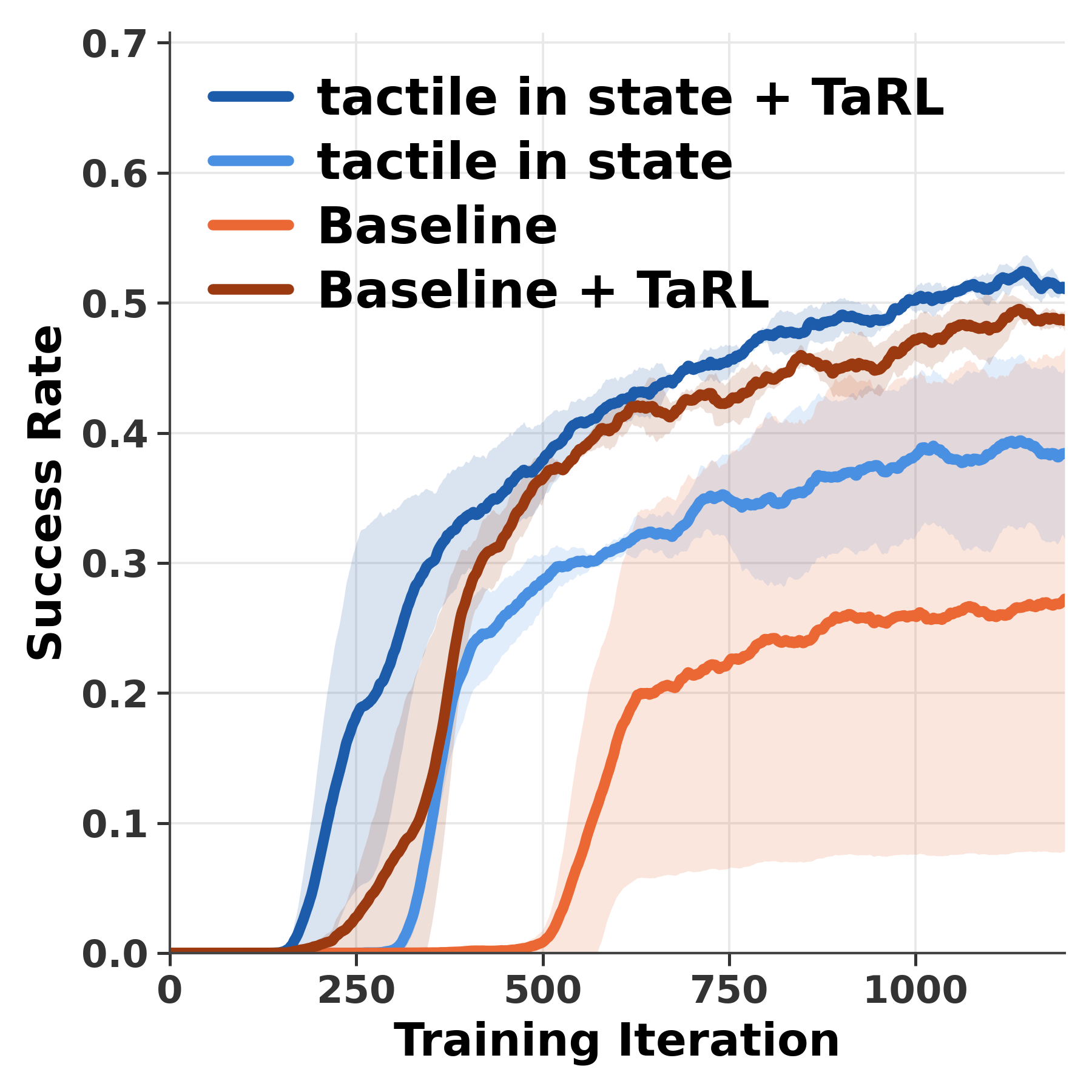}{0.33} &
    \imw{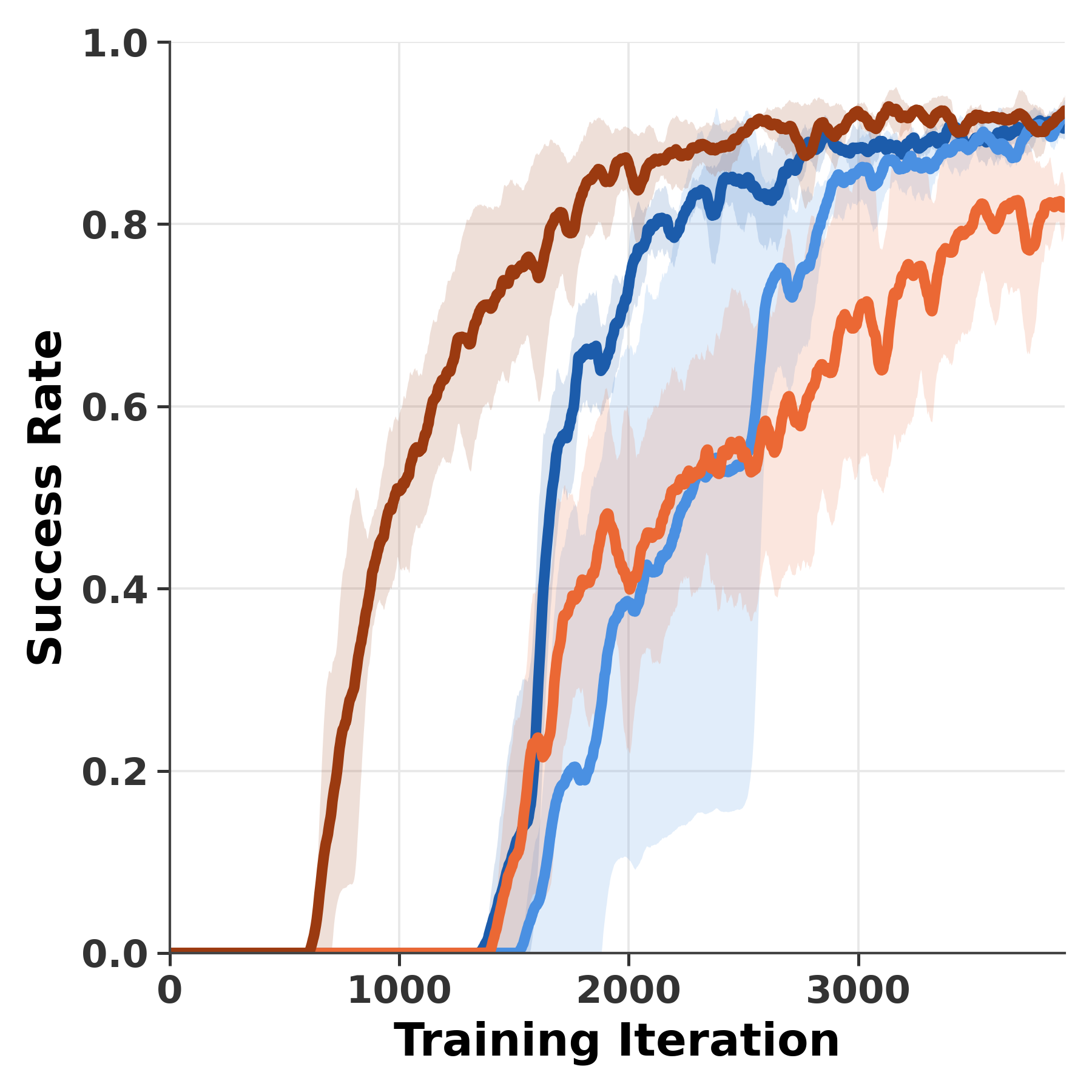}{0.33} & 
    \imw{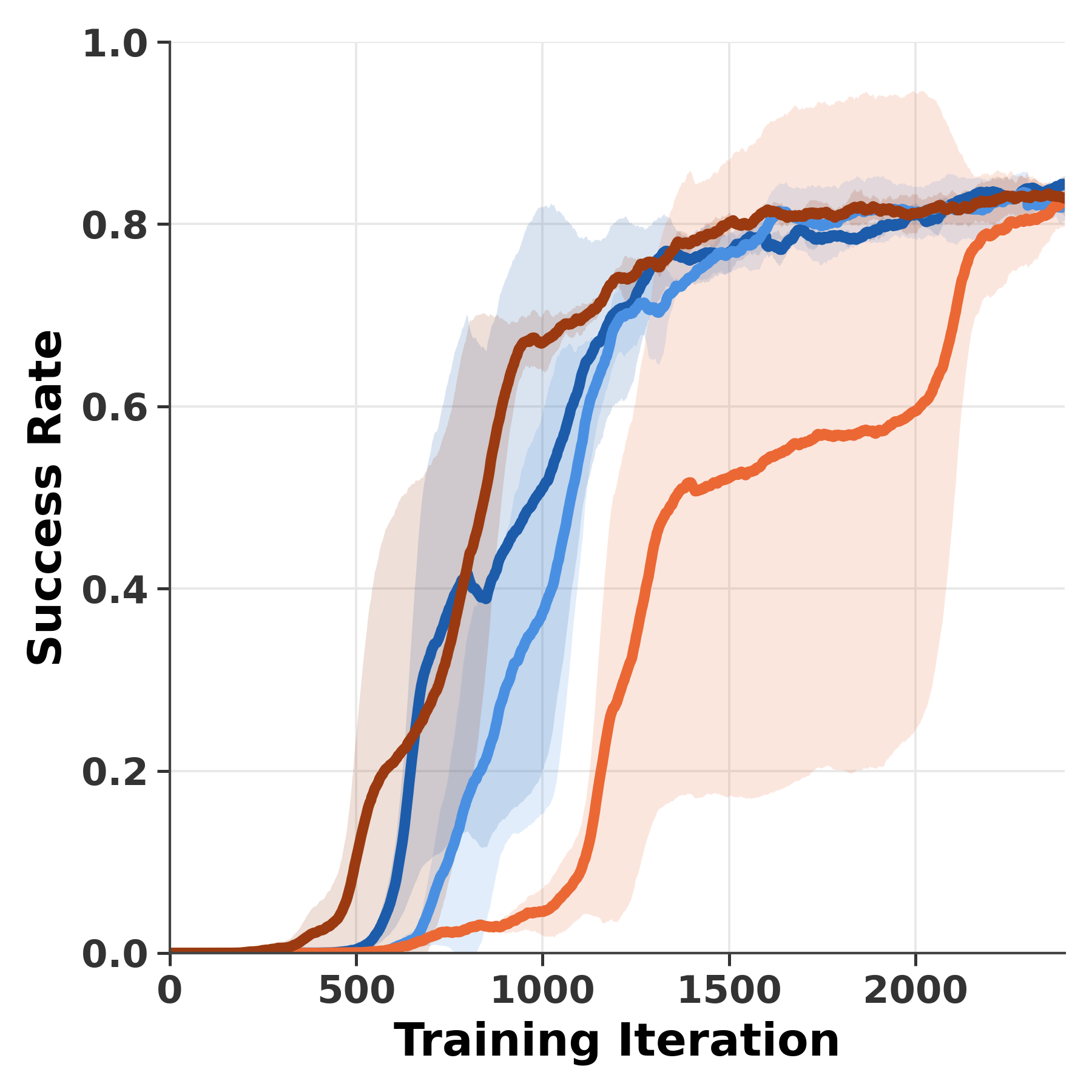}{0.33} \\
    Gear assembly &
    Peg insertion & Nut threading
    }
    \end{adjustbox}
    \caption{
      \textbf{Tactile-based RL efficiency.}  We train a tactile-based RL policy with actors and critics both taking tactile observations as inputs.  For the tactile-based (state-based) approach, we evaluate its downstream RL efficiency 
      \textcolor{blue}{with} (\textcolor{brown}{with}) and \textcolor{cyan}{without} (\textcolor{orange}{without}) \method{} on gear assembly, peg insertion and nut threading tasks in simulation.  Using \method{} enhances training efficiency and task success in both cases.
    }
    \label{fig:TaRL_emb}
    \vspace{-15pt}

\end{figure}

We first study whether tactile rewards naturally emerge from the output of an RL's critic function taking tactile observations as input.  If so, using \method{} as a shaping reward will bring little advantage on the RL learning efficiency, and learning tactile rewards becomes unnecessary.
We conduct experiments on peg insertion, gear assembly and nut threading tasks to verify the conjecture.  We first learn an autoencoder to convert input tactile observations into a latent vector, and concatenate it with the vector of environment states.  We then build a variant of actor and critic function, taking the concatenated vector as input and learning to output actions / values via RL.  We test the RL learning efficiency of this model with and without \method{}.  As shown in \cref{fig:TaRL_emb}, \method{} still brings substantial gain to the learning efficiency and the task success. 

\begin{wrapfigure}{r}{0.45\linewidth}
    \vspace{-12pt}
    \centering
    \includegraphics[width=\linewidth]{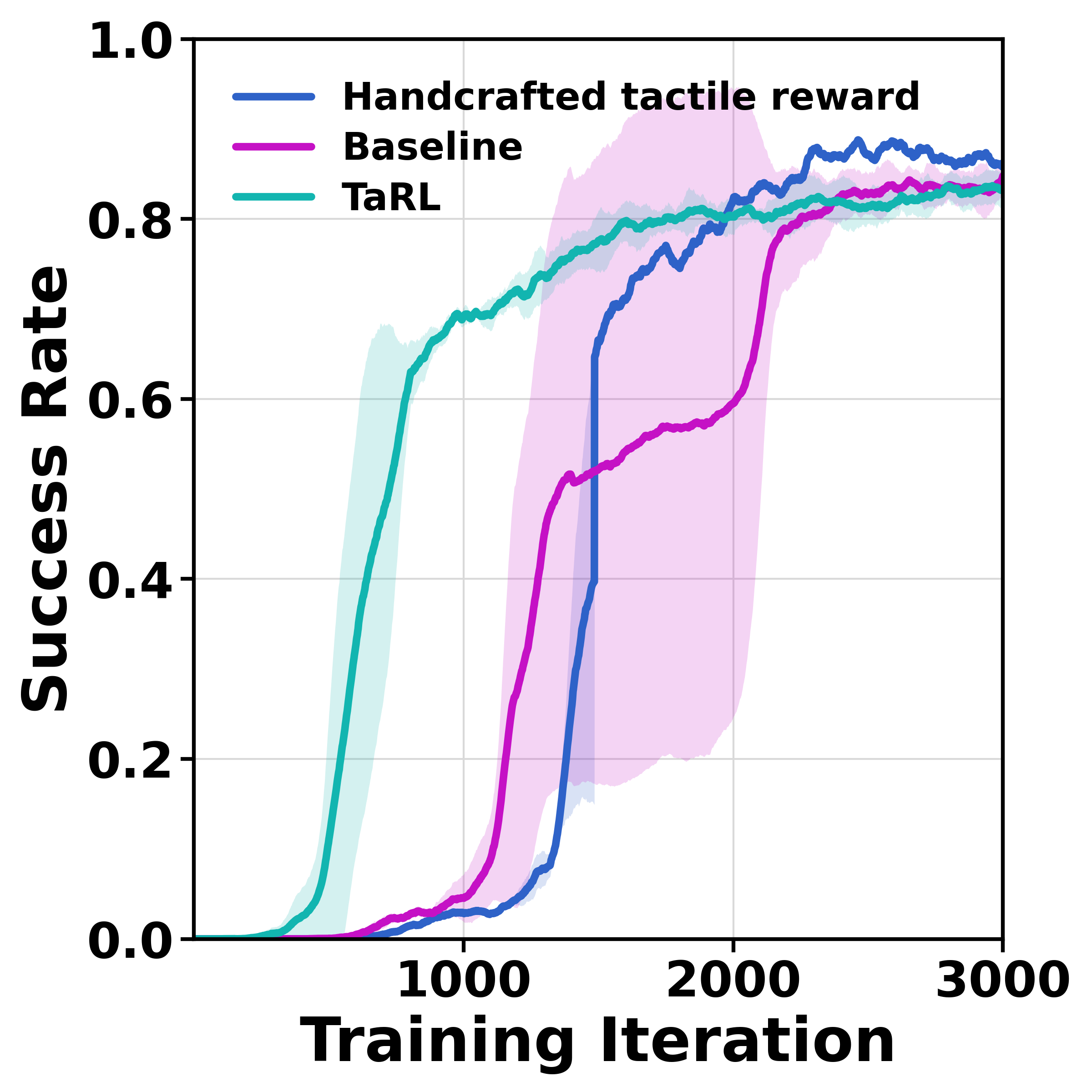}
    \caption{
      \textbf{Comparison of handcrafted tactile reward and \method{}.}     
    }
    \label{fig:heuristic}
    \vspace{-10pt}
\end{wrapfigure}

We next study whether our learned tactile rewards are more effective than handcrafted counterparts.  To manually extract tactile rewards, we devise a tactile-based grasp detector, and design a function that offers reward signals whenever the target object is grasped.  We conduct experiments on nut threading task in simulation, comparing the downstream RL efficiency of \method{} against that of handcrafted tactile rewards.  As shown in \cref{fig:heuristic}, the model optimized with handcrafted tactile rewards demonstrates lower learning efficiency than that with \method{}.  Because the handcrafted reward function fails to distinguish good / bad behaviors once the object is grasped, it offers limited reward signals at different contact phases thereby leading to suboptimal RL performance.   We conclude that useful tactile rewards do not emerge from tactile-conditioned actors and critics, nor from handcrafted reward functions, but via learning from demonstrations.

\subsection{Real-World Experiments}
We validate \method{}'s efficacy of training RL policies for \textit{cube pickup} and \textit{peg insertion} in the real world.  All demonstrations are collected with teleoperation.  For training \method{}, we collect 40 successful and 40 failed trajectories collected at a single object position.  For training policies, we adopt offline RL, collecting 90 successful and 90 failed trajectories at positions different from \method{}'s training data (one new position for peg insertion and two for cube pickup).  For both tasks, we label the fixed dataset with sparse task-completion rewards produced by a human, and dense task-progress rewards produced by \method{}.

\begin{figure}[H]
    \centering
    \vspace{-10pt}
    \begin{adjustbox}{width=\linewidth}
    \tb{@{}ccc@{}}{0.1}{
    \imw{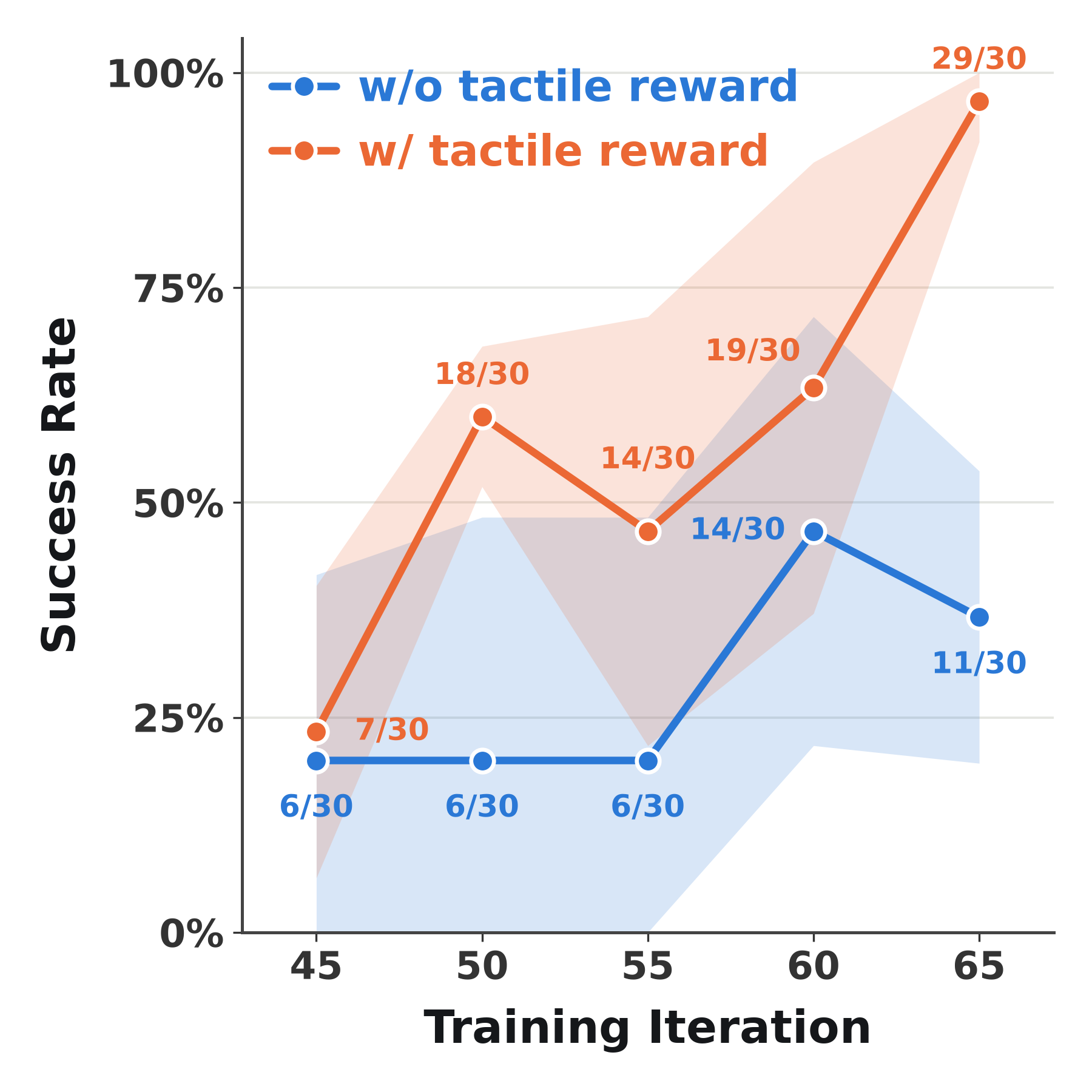}{0.33} &
    \imw{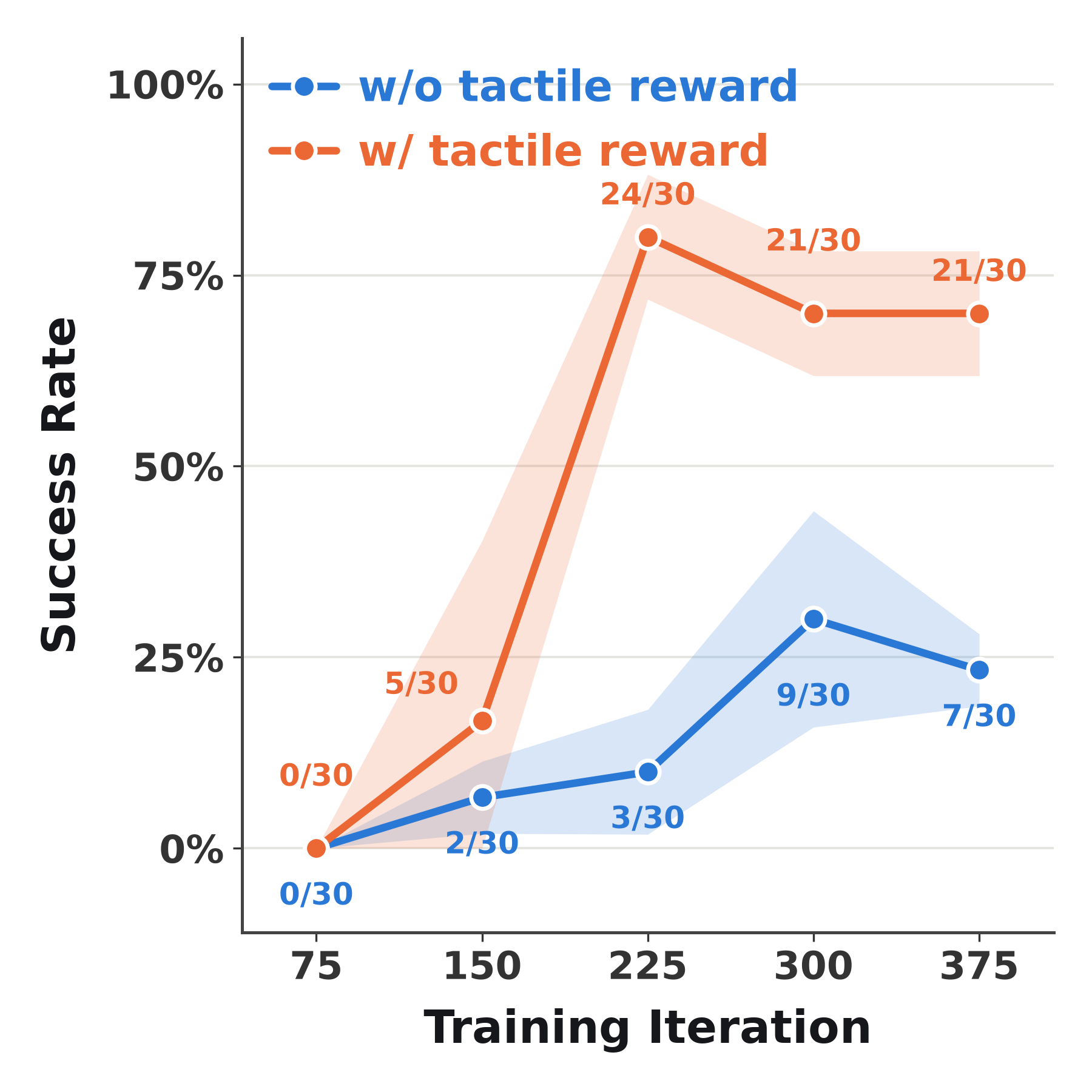}{0.33} &
    \imw{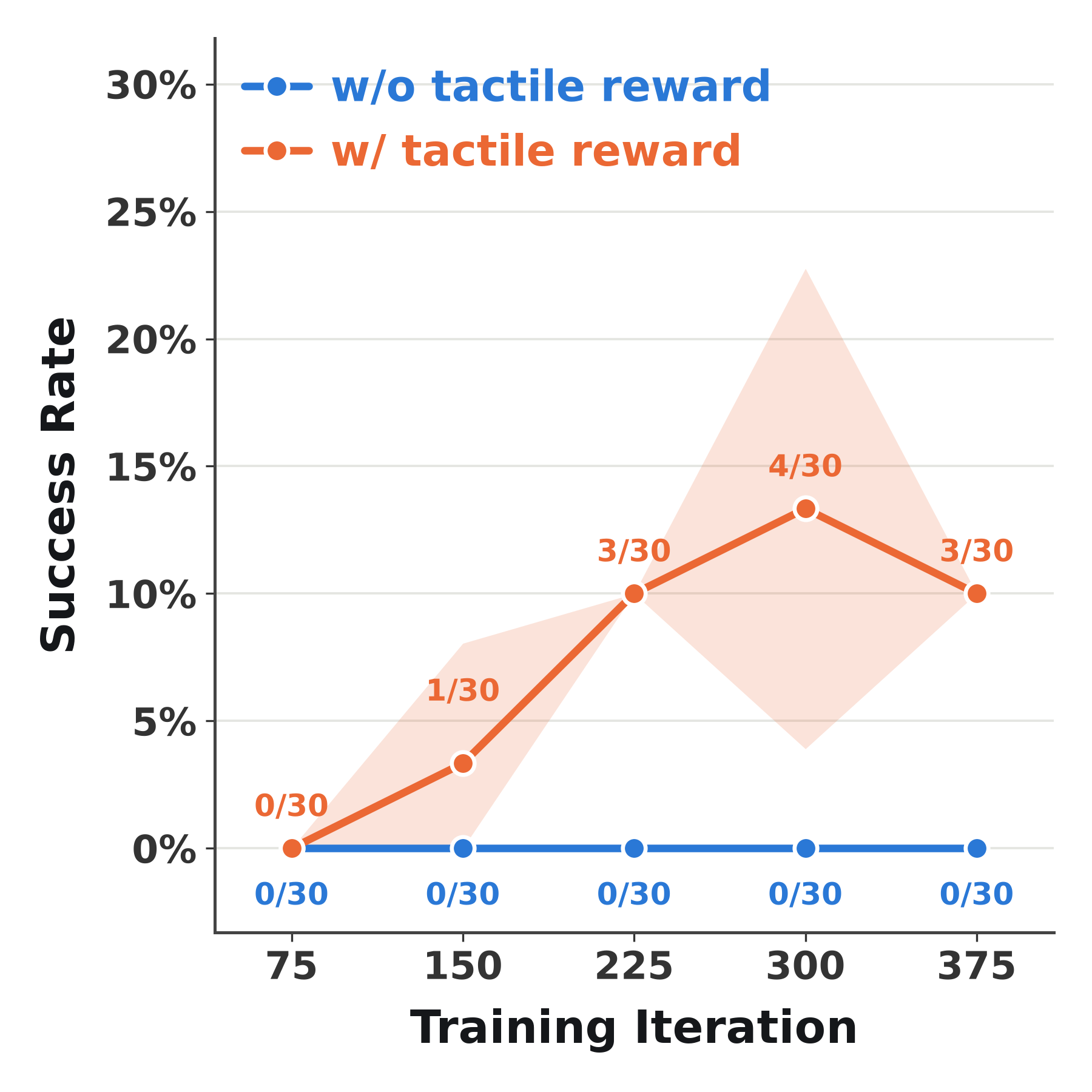}{0.33}  \\
    Cube Pickup & Peg Pickup & Peg Insertion
    }
    \end{adjustbox}
    \caption{\textbf{Real-world RL efficiency.}  We compare downstream RL efficiency \textcolor{orange}{with} and \textcolor{blue}{without} \method{} in the real world.  From left to right, we show the success rate during inference on cube pickup, peg pickup and peg insertion tasks.  Policies optimized with both basic sparse rewards and tactile rewards substantially outperform those with only sparse rewards.
    }
    \label{fig:real_world}
    \vspace{-10pt}
\end{figure}

We train policies with and without tactile rewards using IQL across 3 random seeds, and evaluate the policy's performance with the final task success rate.  Because the peg insertion task includes two goals---picking up the peg and inserting it into the hole, we report the result for each phase separately.  As shown in \cref{fig:real_world}, policies trained with tactile rewards outperform those without tactile rewards, consistently across all tasks.  For picking up the cube, using \method{} increases the task success from 37\% to 97\%.  For picking up the peg and inserting the peg, \method{} achieves an absolute performance gain of 45\% and 10\%.  

% \section{Limitations}
% In this paper, we introduce Tactile Reward Learning (TaRL), a novel framework that learns dense reward functions directly from tactile demonstrations to guide contact-rich manipulation. By relying on tactile deformation maps rather than visual data, TaRL captures critical physical interactions without requiring ground-truth states, visual observations, or action labels. Our evaluations demonstrate that TaRL significantly accelerates downstream reinforcement learning and exhibits robust generalization to unseen object positions, novel instances, and zero-shot sim-to-real transfer.

% A primary limitation of TaRL is that it is currently trained and evaluated on a single task. Future work will focus on collecting diverse tactile data across multiple tasks to improve the model's generalizability.

% \paragraph{Limitations.}
% \paragraph{Conclusion.}

\section{Conclusion}
We presented \method{}, the first framework that learns reward functions from tactile demonstrations. By regressing task progress from sequences of deformation maps, TaRL captures the physical structure of contact-rich manipulation that visual rewards miss. Across four simulated tasks it improves the sample efficiency and final success of downstream RL, generalizes to unseen object positions and to novel object instance without retraining, and outperforms both tactile policy inputs and handcrafted tactile rewards. In the real world, TaRL raises cube pickup success from 37\% to 97\% with offline RL. Our analysis further shows that visual and tactile rewards are informative at different phases of a task and that combining them yields the best policies. We view this as evidence that reward learning for manipulation should be multi-modal, with touch supplying the contact-level signal that vision cannot.

\textbf{Limitations.} \method{} is trained on a single task at a time, so each new task requires its own set of tactile demonstrations and its own reward model. Whether a single multi-modal reward model, trained jointly on vision and touch across a large and diverse task suite, can transfer to new tasks without retraining remains open.

\section{Acknowledgement}
\label{sec:Acknowledgement}

This work was founded and supported by Delta Electronics Inc.


% \input{7_appendix}

\bibliographystyle{IEEEtran}
\bibliography{references}

\end{document}